\documentclass{article}

\PassOptionsToPackage{numbers, compress}{natbib}
\usepackage[final]{neurips_2024}

\usepackage[utf8]{inputenc} % allow utf-8 input
\usepackage[T1]{fontenc}    % use 8-bit T1 fonts
\usepackage{hyperref}       % hyperlinks
\usepackage{url}            % simple URL typesetting
\usepackage{booktabs}       % professional-quality tables
\usepackage{amsfonts}       % blackboard math symbols
\usepackage{nicefrac}       % compact symbols for 1/2, etc.
\usepackage{microtype}      % microtypography

\usepackage{natbib}
\usepackage{amsmath,amsfonts,amsthm,amssymb,multirow}
\usepackage{algorithm, algorithmic}
\usepackage{varwidth}
\usepackage{arydshln}

\usepackage{microtype}
\usepackage{graphicx}
\usepackage{booktabs} % for professional tables
\usepackage[dvipsnames]{xcolor}
\usepackage{subcaption}
\PassOptionsToPackage{hyphens}{url}\usepackage{hyperref}
\usepackage{hyperref}

\DeclareMathOperator*{\minimize}{minimize}

\usepackage{amsmath}
\usepackage{amssymb}
\usepackage{mathtools}
\usepackage{amsthm}

\usepackage[capitalize,noabbrev]{cleveref}

\theoremstyle{plain}
\newtheorem{theorem}{Theorem}[section]

\newtheorem{lemma}[theorem]{Lemma}

\theoremstyle{definition}
\newtheorem{definition}[theorem]{Definition}
\newtheorem{assumption}[theorem]{Assumption}
\theoremstyle{remark}
\newtheorem{remark}[theorem]{Remark}
\newtheorem*{challenges*}{Challenges}
\newtheorem{challenge}{Challenge}
\usepackage{enumitem}
\usepackage{thmtools,thm-restate}

\DeclareMathOperator*{\argmin}{arg\,min}
\DeclareMathOperator*{\amin}{\mathrm{argmin}}
\DeclareMathOperator*{\amax}{\mathrm{argmax}}

\newcommand*{\Scale}[2][4]{\scalebox{#1}{$#2$}}%

\title{Initializing Services in Interactive ML Systems for Diverse Users}

\author{
Avinandan Bose\\ University of Washington\\
avibose@cs.washington.edu \And Mihaela Curmei\\University of California Berkeley\\mcurmei@berkeley.edu \And Daniel L. Jiang\\ University of Washington\\ danji@cs.washington.edu
\And Jamie Morgenstern\\University of Washington\\jamiemmt@cs.washington.edu \And Sarah Dean\\Cornell University\\sdean@cornell.edu \And Lillian J.~Ratliff \\University of Washington\\ratliffl@uw.edu \And Maryam Fazel\\University of Washington\\mfazel@uw.edu 
}

\begin{document}
\maketitle

% \twocolumn[
% \icmltitle{Initializing Services in Interactive ML Systems for Diverse Users}

% It is OKAY to include author information, even for blind
% submissions: the style file will automatically remove it for you
% unless you've provided the [accepted] option to the icml2024
% package.

% List of affiliations: The first argument should be a (short)
% identifier you will use later to specify author affiliations
% Academic affiliations should list Department, University, City, Region, Country
% Industry affiliations should list Company, City, Region, Country

% You can specify symbols, otherwise they are numbered in order.
% Ideally, you should not use this facility. Affiliations will be numbered
% in order of appearance and this is the preferred way.

% this must go after the closing bracket ] following \twocolumn[ ...

% This command actually creates the footnote in the first column
% listing the affiliations and the copyright notice.
% The command takes one argument, which is text to display at the start of the footnote.
% The \icmlEqualContribution command is standard text for equal contribution.
% Remove it (just {}) if you do not need this facility.

%\printAffiliationsAndNotice{}  % leave blank if no need to mention equal contribution
% \printAffiliationsAndNotice{\icmlEqualContribution} % otherwise use the standard text.

\begin{abstract}
% \mf{We should update the abstract to follow the new intro}
% This paper investigates the initialization of multi-learner machine learning (ML) systems \avi{let's not use multi-learner, but rather have a one line description of multiple services + users choosing best?}. \mf{sounds good, yes please take a stab at abstract. Can briefly describe problem (1) also} 
This paper investigates ML systems serving a group of users, with multiple models/services, each aimed at specializing to a sub-group of users. We consider settings where upon deploying a set of services, users choose the one minimizing their personal losses and the learner iteratively learns by interacting with diverse users. Prior research shows that the outcomes of learning dynamics, which comprise both the services' adjustments and users' service selections, hinge significantly on the initialization. However, finding good initializations faces two main challenges: (i) \emph{Bandit %preference 
feedback:} Typically, data on user preferences are not available before deploying services 
and observing user behavior; 
 (ii) \emph{Suboptimal local solutions:} The total loss landscape (i.e., the sum of loss functions across all users and services) is not convex 
 and gradient-based algorithms can get stuck in poor local minima.

We address these challenges with a randomized algorithm to adaptively select a minimal set of users for data collection in order to initialize a set of services. 
 Under mild assumptions on the loss functions, we prove that our initialization 
 leads to a total loss  
  within a factor of the \textit{globally optimal total loss 
 with complete user preference data}, and this factor scales logarithmically in the number of services. This result is a generalization of the well-known $k$-means++ guarantee to a broad problem class, which is also of independent interest.
 The theory is complemented by experiments on real as well as semi-synthetic datasets.
\end{abstract}

%\mf{Also the term 'unknown preference' confused the reviewers before, 'unknown' is not the right term and we use it very very often, let's be clear what we mean, we can refer to it as bandit feedback (plus the other query type that we discuss later). I think we updated intro but rest of the paper and appendices need a check.}\avi{Have a look at abstract now. Changed unknown preferences to bandit preference feedback in abstract and intro.}

% }
\section{Introduction}\label{sec:intro}
We consider a setting where a {provider} wants to design $k$ {services} for $n$ {users} with diverse preferences. Each service %is aimed at specializing in a subgroup of users and 
uses a model parameterized by a vector $\theta \in \mathbb{R}^d$ to predict users' preferences, and 
users pick a service 
that yields the smallest loss for them.
The loss incurred by user $i$ when choosing a service parameterized by $\theta$ is denoted by $\mathcal{L}_i(\theta, \phi_i)$, where $\phi_i \in \mathbb{R}^d$  
parameterizes the user’s preference. 
We want to design $k$ services by minimizing the sum of all losses; 
i.e., an optimization problem of the form:
%We consider the setting where a {provider} wants to design $k$ {services} for $n$ {users} with (unknown) diverse preferences, such that each service specializes to a subgroup of users. Each service uses a model parameterized by a vector $\theta \in \mathbb{R}^d$, which predicts users' preferences.
%A user is more likely to use a service that yields higher utility by predicting their preference more accurately---i.e., corresponding to lower prediction loss. 
%
%Let the loss incurred by user $i$ upon choosing some service parameterized by $\theta \in \mathbb{R}^d$ be denoted by $\mathcal{L}_i(\theta, \phi_i)$, where $\phi_i \in \mathbb{R}^d$ represents this user's %(unknown) preference parameter and $\mathcal{L}_i : \mathbb{R}^d \times \mathbb{R}^d \rightarrow \mathbb{R}_{+}$ is their loss function. We consider designing the $k$ services (i.e., choosing parameters $\theta_1,\ldots\theta_k$ by minimizing the sum of all the losses, i.e., an optimization problem with the following form:
%
\begin{align}\label{eq:opt_prob}
\minimize_{\theta_1,\ldots, \theta_k \in \mathbb{R}^d} \;\; 
& \sum_{i=1}^{n} \min \{\mathcal{L}_i(\theta_1, \phi_i),\ldots
\mathcal{L}_i(\theta_k, \phi_i)\}.
\end{align}
This problem formulation is broad and includes the classical $k$-means clustering problem \cite{lloyd1982least} (where $\mathcal{L}_i$ are Euclidean distances and the inner `min' selects the closest among $k$ centroids), mixed-linear regression \citep{zhong2016mixed}, generalized principal component analysis (GPCA) or subspace clustering \cite{vidal2005GPCA}, in addition to our new, motivating problem of designing $k$ services for $n$ users. Even if the losses $\mathcal{L}_i$ are convex in $\theta$, this objective is generally not convex (even in the special case of the $k$-means problem). 

Our goal is to find a local minimum of this optimization problem with an approximation ratio (a worst-case guarantee on the achieved total loss with respect to the global optimum) under suitable yet broad assumptions on $\mathcal{L}_i$.
%Achieving this relies on good initialization and the use of problem structure. 
Further, an important limitation in many practical settings is that the provider/designer has only \emph{bandit feedback} (zeroth-order oracle access) to the loss functions (i.e., the designer doesn't know the function $\mathcal{L}_i(\cdot, \phi_i)$, but can only evaluate its value for some $\theta$ corresponding to the service chosen by the user among the ones deployed), which further complicates solution methods, compared to the classical cases of clustering and facility location problems \citep{cornuejols1983uncapicitated} which typically assume full information. 

In this paper, we seek a novel and effective initialization scheme that vastly extends the celebrated $k$-means++ algorithm and its analysis \cite{arthur2007k}. This scheme should retain the simplicity and ease of implementation of the original algorithm, yet be able to (1) handle general loss families (assumptions on $\mathcal{L}_i$ are discussed in Section~\ref{sec:preliminaries}), (2) provide a tight, instance-dependent approximation ratio (details in Section~\ref{sec:main_results}), 
and (3) handle realistic information limitations such as access only to (noisy) bandit feedback, in a sample-efficient manner. 
Next, we describe in more detail the important use case of initialization of services for diverse users in multi-service ML systems (yet as noted above, our main result has other applications as well, and can be of independent interest). 
 
%\textbf{Bandit/zeroth-order access to loss functions.} An important limitation in practice is that the provider only has query access to the value of the users' loss functions, that is, only function-value evaluation or bandit feedback from the loss functions is available (see Section~\ref{sec:main_results} for more details). We also extend this to the case where the feedback is noisy (see Section~\ref{sec:linear}). 

%\mf{Please take a pass at editing the rest of the intro,  needs to be edited to connect more clearly with the motivation above. Then need to update the abstract too.}

\textbf{Motivation.} In a variety of contexts such as federated learning \citep{li2020federated}, crowd-sourcing \citep{steinhardt2016avoiding} and online recommendation systems \citep{song2014online}, data about user preferences is acquired through iterative interactions. This data is then used to improve the model and serve the individual needs of users. 
Given that users' preferences are typically heterogeneous, recent works demonstrate that using multiple specialized models can be more effective than the one-size-fits-all approach of employing a single large shared model, 
%Model specialization is found to be effective 
e.g., for clustered federated learning \cite{mansour2020three,sattler2020clustered,ghosh2020efficient}, %\citep{dwork2018decoupled},  
 meta learning \citep{kong2020meta, bose2024offline}, 
fine-tuning for specific groups of users or tasks \citep{Chua2021finetuning,sun2021towards}, and
in the context of fair classifiers \citep{ustun2019fairness}. 
Here we tackle the crucial yet under-explored phase of \emph{initializing services} in ML systems that learn interactively from diverse users.
%, each with distinct data distributions. 
The initialization process is crucial as it sets the stage for how effectively these systems can adapt and specialize with future user interactions. Once initialized, the services interact with users, who, in turn, choose among services based on their loss. These ``learning dynamics'' typically lead to the specialization of services to groups of users \citep{ginart2021competing, dean2022multi}, and  
\cite{dean2022multi} shows experimentally that the overall social welfare achieved by the services depends on the initialization of the learning dynamics. 
We note that in the context of problem \eqref{eq:opt_prob}, the learning dynamics in \cite{dean2022multi} can be seen as updates of an alternating minimization algorithm, iteratively updating users' choices (the inner minimization) and services' parameters (update to each $\theta_j$).  %\cite{dean2022multi} shows a broad class of learning dynamics converge to a local minimum of the loss, whose value depends on the initialization. (this is analogous to running Lloyd's algorithm \cite{lloyd1982least} after $k$-means++ initialization). 
Our goal is to initialize a set of services to minimize the sum of losses for all users (or equivalently, maximize total welfare), tackling the following challenges:
\begin{itemize}[itemsep=-2pt, topsep=-2pt, left=0pt]
    \item \textbf{Bandit loss feedback:} 
    In practice, %the presence of 
    offering a service often precedes data collection. %, rather than the other way around. 
    Specifically, in contexts like online recommendations, it is usually not feasible to gather user preference data (knowledge about $\phi_i$ and evaluations of $\mathcal{L}_i(\cdot, \phi_i)$ at various different parameter values in problem~\eqref{eq:opt_prob}) without first deploying the services parameterized by $\{\theta_1, \ldots, \theta_k\}$ and observing user interactions. This means that data collection is inherently conditional on the existence of services, challenging the conventional ``data-first, model-second'' paradigm.
    \item \textbf{Suboptimal local solutions:} 
    % \mf{make this clear: which function in which variables} 
    % Despite individual user losses and service losses \mf{maybe not say this since we didn't define service loss above} being convex, the total loss across all users becomes non-convex with respect to all the parameters of the set of services due to the fact. 
    Since users select the service with the lowest loss ($\min \{\mathcal{L}_i(\theta_1, \phi_i),\ldots
\mathcal{L}_i(\theta_k, \phi_i)\}$ in problem~\eqref{eq:opt_prob}), minimizing over the parameters $\{\theta_1, \ldots, \theta_k\}$ leads to a  nonconvex problem 
in general, as mentioned earlier.  %(without very restrictive assumptions on $\mathcal{L}_i(\cdot,\phi_i)$). 
    %The total loss (i.e., the sum of losses across all users) is  non-convex jointly in both sets of variables for users and services even if the individual user losses on a single service are convex
Gradient-based 
learning dynamics can get stuck in local minima where the total loss can be significantly worse than the globally minimum loss. Thus, the outcomes of learning dynamics are heavily affected by initialization of service parameters.
\end{itemize}

%\ljr{suggestion: if we have space, make this a subsection. stylistically its weird to have only one subsection...i.e. related work.}

\textbf{Contributions.} The following summarizes the contributions of this paper.
    \begin{itemize}[itemsep=-2pt, topsep=-2pt, left=0pt]
    \item  We design a computationally and statistically efficient algorithm for \emph{initialization of services} prior to learning dynamics. The algorithm works by adaptively selecting a small number of users to collect data from (via queries of their loss function) in order to initialize the set of services. 
    % \mf{actually it is probably better to move these sentences to Contributions as a first bullet point that describes the algorithm: we propose an algorithm that...} \mf{also in the contributions we should emphasize that `gathering information' is based on low-information queries (or the terms for this we use later)... it is not coming across.}
    \item We establish an \emph{approximation ratio} for the designed algorithm: the expected total loss achieved by the algorithm {right after initialization} is within a factor of the \textit{globally optimal total loss in the presence of complete user preference data}, and this factor scales  logarithmically in the number of services. Furthermore this bound is tight, and recovers the known k-means++ approximation ratio as a special case  (cf.~Section~\ref{sec:main_results}).  
    \item When users belong to a set of demographic groups, it is desirable that the services do not result in unfavorable outcomes towards certain demographics (e.g., based on gender or racial groups). One fair objective is to minimize the maximum average loss of users across different groups. We provide an \textit{approximation ratio} for this fair objective that scales logarithmically in the number of services (cf.~Section~\ref{sec:main_results}).  
    \item In the context of linear prediction models, we study the problem of \emph{generalizing} to users the provider has not interacted with before (cf.~Section \ref{sec:linear}).     
    \item We empirically demonstrate the strengths of our initialization scheme via experiments on a prediction task using 2021 US Census data, and online movie recommendation task using the Movielens10M dataset (cf.~Section \ref{sec:experiments}).
\end{itemize}

\subsection{Related Work}

% \mf{this paragraph can move to appendix} \textbf{Retention.} User retention in machine learning systems is closely related to the decision dynamics between the provider and users studied in the single service setting by \citet{hashimoto2018fairness, zhang2019group} and multiple services setting by \citet{dean2022multi, ginart2021competing}. In settings with multiple
% sub-populations of users of different types, the question of retention has been explored in parallel
% with the issue of fairness. These works typically focus on the stability of the dynamics at equilibrium. However, the final outcome is heavily influenced by the initial data the provider has and the initial configuration (partitioning) of the users across the offered services. Our work addresses \textit{the impact of initialization on the final outcome with theoretical guarantees.}

\textbf{Multiple Model Specialization.} 
% We consider the inherent heterogeneity in user preferences, especially 
In distributed learning, where data sources are users' personal devices,
% . For applications such as recommendation systems and personalized advertising
utilizing multiple specialized models, where users are grouped into clusters representing interests, can yield improved predictions and outcomes. 
For instance, in recommendation systems these clusters could represent users interested in different movie genres, or different combination of features (see Appendix~\ref{appendix:example} for a concrete example on Netflix recommendation clusters). This approach has been adopted %in recent federated learning formulations such as 
recently in clustered federated learning \citep{sattler2020clustered, mansour2020three, ghosh2020efficient} and online interactive learning \citep{narang2022online}, facility location problems \cite{bose2022scalable}, where users \emph{choose} models/services and for which they provide updates. 
%Prior theoretical study typically assume users belong to clusters that are well separated, which our results do not require. \mf{this sentence is probably better in the clustering paragraph}

% \mf{more citations in this paragraph, or shorten the paragraph? As a related work section it only cites 3 papers} We
% consider the inherent heterogeneity in user preferences, especially prominent in distributed learning where data sources are users' personal devices. For applications such as recommendation systems and personalized advertising, utilizing multiple specialized models, where users are grouped into clusters representing interests, can yield improved predictions and outcomes. 
% For instance, in movie recommendation systems these clusters could represent users interested in different movie genres, or different combination of features. This approach has been adopted %in recent federated learning formulations such as 
% recently in clustered federated learning \citep{sattler2020clustered, mansour2020three} and is integral to the federated learning model of \citep{ghosh2020efficient}, where users \emph{choose} models for which to provide updates. 
% Prior theoretical study typically assume users belong to clusters that are well separated, which our results do not require. 

\textbf{Clustering.} 
Multiple model specialization leads to clustering the users into groups and centering a specialized model on each group. We provide a brief review of the $k$-means clustering problem and establish the connection to specialization. The $k$-means clustering problem is one of the most commonly encountered unsupervised learning problems. Given a set of known $n$ points in Euclidean space, the goal is to partition them into $k$ clusters (each characterized by a center), such that the sum of square of distances to their closest center is minimized. \citet{dasgupta2008hardness} and \citet{aloise2009np} showed that the $k$-means problem is NP-Hard. 
The most popular heuristic for $k$-means is Lloyd's algorithm \citep{lloyd1982least}, 
%often referred to as the $k$-means algorithm, 
which proceeds by randomly initializing $k$ centers and then uses iterative updates to find a locally optimal $k$-means clustering, which can be arbitrarily bad compared to the globally optimal clustering. The performance of the $k$-means algorithm relies crucially on the initialization. \citet{arthur2007k} and \citet{ostrovsky2013effectiveness} proposed an elegant polynomial time algorithm for initializing centers, known as $k$-means++. \citet{arthur2007k} proved that the expected cost of the initial clustering obtained by $k$-means++ is at most $8(2 + \log k)$ times the cost of optimal $k$-means clustering. 
Our work generalizes the analysis of \citet{arthur2007k} to the setting where \textit{user's preferences are represented as unknown points and the loss functions are unknown with only bandit access, not necessarily identical, and general as long they satisfy Assumptions~\ref{ass:unique} and \ref{ass:triangle}}, with important examples given in Appendix \ref{sec:examples}. 

For a detailed discussion on more related works please see Appendix~\ref{sec:additional_related_works}.
%
% \textcolor{blue}{Algorithms used for GMMs, like expectation maximization  can only guarantee local convergence (e.g., \citep{xu1996convergence}). \citep{jin2016local} showed that GMMs can have arbitrarily bad local optima even in special case of well separated Gaussians.} \mf{let's remove this discussion of EM algorithms, they're different and discussing them may confuse the reader. Just reference to the GMM model above should be enough.}

\textbf{Notation and Terminology.} 
For a symmetric matrix $\mathbf{A}$ and any vector $x \in \mathbb{R}^d$, we denote its Mahalanobis norm by $\|x\|_\mathbf{A}=\sqrt{x^\top \mathbf{A} x}$.
The generalized eigenvalues  for a pair of symmetric matrices $\mathbf{A}$ and $\mathbf{B}$ are denoted by $\lambda(\mathbf{A}, \mathbf{B})$, defined as the solutions of $\lambda$ for the generalized eigenvalue problems $\mathbf{A} v = \lambda \mathbf{B} v$ 
\citep{ghojogh2019eigenvalue,forstner2003metric}. Specifically we use  $\lambda_{\min}(\mathbf{A}, \mathbf{B})$ to denote minimum generalized eigenvalue for the matrix pair $\mathbf{A}, \mathbf{B}$. 
The loss for a user $i$ given service $\theta \in \mathbb{R}^d$ is denoted by $\mathcal{L}_i(\theta,\phi_i)$ where $\phi_i$ parameterizes the user's preference.
For a set of users $\mathcal{A}$ (e.g., $\mathcal{A} = [n]$ denotes a set of $n$ users) and a set of services $\Theta = \{\theta_1, \ldots, \theta_k\} \subset \mathbb{R}^d$, the total loss is defined as $\mathcal{L}(\Theta, \mathcal{A}) = \sum_{i \in \mathcal{A}}\min_{j \in [k]} \mathcal{L}_i(\theta_j, \phi_i)$.

\section{Problem Setup}\label{sec:preliminaries}
We make the following assumptions about the functional form of $\mathcal{L}_i$, and state several examples of function classes satisfying these properties. Note that %we ({the algorithm designer}) 
the designer/provider doesn't need to know the functional form of $\mathcal{L}_i$, and knowledge about $\mathcal{L}_i$ is obtained through bandit feedback via observing the scalar values $\mathcal{L}_i(\theta, \phi_i)$ for different $\theta$, where $\theta$ parameterizes the services. 

\begin{assumption}[Unique Minimizer]\label{ass:unique}
The loss function satisfies the following equivalence: $\mathcal{L}_i(\theta, \phi_i) = 0 \ \Longleftrightarrow\ \theta = \phi_i$.
\end{assumption}
This assumption implies that unless all users have identical preference parameters, there doesn't exist a single service parameter $\theta$ that simultaneously minimizes every user's loss. Thus providing multiple services (multiple $\theta$'s) where the users choose the one best for them is strictly better than one service for all users.
%a single service for every user. 
%\mf{This is a bit confusing. There could be different losses that happen to have the same minimizer. Should make it more clear. Also the last phrase may be confused as referring to $n$ services, could say instead `a single service for all users'.}
% \mf{I edited this, please check}
%
\begin{assumption}[Approximate Triangle Inequalities]\label{ass:triangle} For a pair of users $i,j$ there exists a finite constant $c_{ij} > 0$ such that for all $ \theta \in \mathbb{R}^d$ the following hold:
    \begin{enumerate}[label={{\it (\roman*)}}, topsep=0pt, itemsep=-2pt]
        \item $c_{ij} \mathcal{L}_i(\theta, \phi_i) \leq \mathcal{L}_j(\theta, \phi_j) + \mathcal{L}_j(\phi_i, \phi_j)$.
        \item $c_{ij} \mathcal{L}_i(\phi_j, \phi_i) \leq \mathcal{L}_j(\theta, \phi_j) + \mathcal{L}_i(\theta, \phi_i)$.
    \end{enumerate}
\end{assumption}
Here $c_{ij}$ (equal to $c_{ji}$) captures the alignment between the preference parameters and loss geometries of two users. Lower values of $c_{ij}$ indicate less similarity. 
Item $(i)$ implies that the loss for user $i$ on any service $\theta \in \mathbb{R}^d$ is no worse than (up to a constant factor) the sum of (a) loss of another user $j$ on using the same service, and (b) the loss of user $j$ if they were to use user $i$'s preference parameter. The latter term (b) can be seen as measuring the similarity between the users' preferences.

For condition $(ii)$ to hold for all $\theta \in \mathbb{R}^d$, it must also hold for the service that minimizes the sum of losses of both users using the same service. Suppose that users $i$ and $j$ were to exchange preference parameters. %
Then their loss would be no worse than (up to a constant factor) their minimum total loss, i.e., \[c_{ij} \mathcal{L}_i(\phi_j, \phi_i) \leq \min_{\theta \in \mathbb{R}^d} \left(\mathcal{L}_j(\theta, \phi_j) + \mathcal{L}_i(\theta, \phi_i)\right).\]
% \[\mathcal{L}_i(\phi_j, \phi_i) \leq \frac{1}{c_{ij}}\min_{\theta \in \mathbb{R}^d} \left(\mathcal{L}_j(\theta, \phi_j) + \mathcal{L}_i(\theta, \phi_i)\right).\]

Some examples of loss functions and the corresponding constants that satisfy these assumptions include the following (see Appendix~\ref{sec:examples} for additional examples and derivations):
\begin{itemize}[itemsep=-2pt, topsep=-2pt, left=0pt]
\item Squared error loss for linear predictors (cf.~Section~\ref{sec:linear}).
\item The Huber loss on the prediction error:
    \begin{equation*}
    \mathcal{L}_i(\theta, \phi_i) =
    \begin{cases}
        \frac{1}{2} \|\theta - \phi_i\|^2, & \text{if } \|\theta - \phi_i\| \leq \delta \\
         \delta(\|\theta - \phi_i\| - \frac{1}{2}\delta), & \text{otherwise.} 
    \end{cases}
\end{equation*}
This loss is used typically in robust estimation tasks.
Here $\|\cdot\|$ could be any norm, and we show $c_{ij} = 1/3$.
\item The normalized cosine distance: $\mathcal{L}_i(\theta, \phi_i) = 1 - \theta^\top \phi_i$ where $\|\theta\|_2 = \|\phi_i\|_2 = 1$, with $c_{ij} = \frac{1}{2}$. This is commonly used as a similarity measure in natural language processing applications, for example finding similarity between two documents.
\item The Mahalanobis distance: $\mathcal{L}_i(\theta, \phi_i) = \|\theta - \phi_i\|_{\Sigma_i}$. Different users can have different $\Sigma_i$ capturing their diverse loss variation, as long as $\Sigma_i$ is full rank. Here  $c_{ij} = \min\{\lambda_{\rm min}(\Sigma_i, \Sigma_j), \lambda_{\rm min}(\Sigma_j, \Sigma_i)\}$.
\item Any distance metric: %$\mathcal{L}_i(\cdot, \cdot)$. 
This naturally follows from triangle inequality, hence $c_{ij} = 1$. 
% (Note that the examples above are also all distances, but when focusing on their special form we obtain better estimates for the $c_{ij}$.) \avi{This is an incorrect statement, so removed.}
% \mf{for added clarification. We can move this to a footnote if it looks better.}
\item Any arbitrary function $\mathcal{L}_i(\theta,\phi_i)$ that is $L_i$-Lipschitz and $\mu_i$-strongly convex in $\theta$ with $c_{ij} = \min(\mu_i, \mu_j) / \max(L_i, L_j)$.
% \mf{should we also comment that taking the worst-case $L/\mu$ for a set of strongly convex losses is a simplya  special case, and in our examples we've worked out tighter instance-based values---and refer to appendix?}\avi{We should mention arbitrary lipschitz and strongly convex functions satisfy this property with bound $L/\mu$. However, knowing the structure of the losses, leads to working out better bounds as in the examples above.} \mf{yes, or if each has $L_i, \mu_i$ then worst case over $i$ can be used, and same argument---we can do much better in the specific examples by working out the the constants. Add what you think is best and I'll edit.}
\end{itemize}

\textbf{Objective.} Suppose the users have access to $k$ services parameterized by $\Theta = \{\theta_1, \ldots, \theta_k\} \subset \mathbb{R}^d$. Then, each user $i$ selects service $\theta_l$ that minimizes their loss, i.e. $\mathcal{L}_i(\Theta, \phi_i) = \min_{l \in [k]} \mathcal{L}_i(\theta_l, \phi_i)$.

As discussed earlier, our goal is to design $\Theta$, such that the sum of losses across users and services is minimized. We define the objective as follows:
\begin{align}\label{eq:obj}
    \Scale[0.9]{\mathcal{L}(\Theta, [n]) = \sum_{i \in [n]} \min_{j \in [k]} \mathcal{L}_i(\theta_j, \phi_i) = \sum_{i \in [n]} \mathcal{L}_i(\Theta, \phi_i).}
\end{align}
\begin{definition}\label{def:clustering}
    Define the unknown optimal set of $k$ services that minimizes the objective to be
\[
    \Scale[0.9]{\Theta_{\rm OPT} := \amin_{|\Theta| = k} \mathcal{L}(\Theta, [n]).}
\]
Specifically, $\Theta_{\rm OPT}$ defines a ``clustering'', meaning a partitioning of the $n$ users into $k$ clusters. 
The cluster $\mathcal{B}_m$ is the set of all users that prefer the service $\theta_m$ among all the services in the optimal set $\Theta_{\rm OPT}$. In other words, $\mathcal{B}_m$ is defined as the set of all points such that $\mathcal{B}_m = \{i \in [n] \;|\; \theta_m = \argmin_{\theta_l \in \Theta_{\rm OPT}} \mathcal{L}_i(\theta_l, \phi_i)\}$. If multiple services are equally preferred by a subpopulation, the ties are broken arbitrarily. The resulting set of clusters is denoted by $\mathcal{C}(\Theta_{\rm OPT}) = \{\mathcal{B}_1, \ldots, \mathcal{B}_k\}$.
\end{definition}

The are several statistical and computational challenges to this problem.
\begin{challenge}
    Since preferences $\{\phi_i\}_{i \in [n]}$ and loss functions $\{\mathcal{L}_i\}_{i \in [n]}$ are unknown and the provider only has zeroth order or bandit feedback access, estimating the objective function $\mathcal{L}(\Theta, [n])$ usually needs a lot of data collected uniformly across the users. This large amount of data is needed before services can be deployed, yet as stated earlier, we are in the situation where we have no data until we deploy services and observe user interactions. Our limited access to user information (via limited queries) makes our setting challenging.  
    % \mf{added this for more emphasis}
\end{challenge}

\begin{challenge}
%    Even if $\mathcal{L}(\Theta, [n])$ was fully known, the loss function is non-convex and iterative minimization approaches from a random initialization are susceptible to getting stuck in arbitrarily poor local optima. Alternatively, computing the optimal clustering first, and then finding the best service for each cluster (convex objective) is NP-Hard. \mf{We are correct that the problem is NP-hard even with full information, but I guess here the reviewer that talked about k-mediods could say this hardness is not too bad: with full information, we can get a constant approx ratio. I rephrased a bit.} 
%In addition to the first challenge, 
The loss function is non-convex and iterative minimization approaches from a random initialization are susceptible to getting stuck in arbitrarily poor local optima. This means computing the optimal clustering first and then finding the best service for each cluster is NP-Hard.  
\end{challenge}
\begin{challenge}
    We do \textit{not} assume any data separability conditions, for example user preference parameters are  drawn from $k$ well separated distributions. Thus we are unable to exploit underlying structure to reduce sample complexity.
\end{challenge}

Despite the challenges, in Section \ref{sec:main_results}, we propose an algorithm that is both statistically and computationally efficient, and %
admits an approximation ratio with respect to the \emph{globally optimal value}, i.e., $\mathcal{L}(\Theta_{\rm OPT}, [n])$.

% \mf{For the Mahalanobis case, how about adding some of the remarks from the previous version of the paper that gave intuition about loss heterogeneity, maybe even the figure showing initialization with different ellipsoidal losses? It can be in the appendix (since now our main contribution and focus is low-information/lack of data---more important and stronger contribution---and the heterogeneous extension to kmeans++ is just a side contribution).}
\begin{algorithm}[t!]
\caption{AcQUIre- Adaptively Querying Users for Initialization}
% \vskip -0.3in
\begin{algorithmic}[1]
\STATE \textbf{Input:} Set of users $[n]$, number of services $k$.
\STATE Choose a user $i$ uniformly randomly from $[n]$.
\STATE Query user $i$'s preference $\phi_i$, set the first service $\Theta_1 = \phi_i$.
\FOR{$t \in \{2, \ldots, k\}$}
        \STATE \textbf{User behavior:} Collect user losses on existing services $\Theta_{t-1}$ : \{$\mathcal{L}_i(\Theta_{t-1}, \phi_i)\}_{i \in [n]}$.
        \STATE \textbf{User Selection:} Sample $l$ from $[n]$ with probability $P(l = i) \propto \mathcal{L}_i(\Theta_{t-1}, \phi_i)$.
        \STATE \textbf{New service:} Query user $l$'s preference $\phi_l$.
        \STATE $\Theta_t = \Theta_{t-1} \cup \phi_l$.
\ENDFOR
\STATE \textbf{Return} $\Theta_k$ 
\end{algorithmic}
\label{alg:initialize}
\end{algorithm}

\section{Algorithm \& Main Results}\label{sec:main_results}
In this section, we present our initialization algorithm with guarantees, Algorithm~\ref{alg:initialize}, and describe how the steps of the algorithm arise naturally in the interactive systems under consideration. Since collecting data uniformly across all the $n$ users can be prohibitively expensive, our goal is to
get data from a minimal number of users.

Each iteration of the loop in the algorithm adds a service sequentially and the loop terminates when there are $k$ services, where $k$ is a predetermined parameter for the algorithm. 
We focus on the loop (lines 4-8) in Algorithm~\ref{alg:initialize}. 

Suppose at time $t-1$, the set $\Theta_{t-1}$ is the set of current $t-1$ services. Then, at time step $t$ the following steps take place.
\begin{itemize}[itemsep=-3pt, topsep=-4pt,left=0pt]
    \item \textbf{User behavior (line 5). } Given the list of services $\Theta_{t-1}$ users are assumed to choose the best service that minimizes user loss. %Finally, 
    Users report their losses with respect to the service they choose from the set of existing services $\{\mathcal{L}_i(\Theta_{t-1}, \phi_i)\}_{i \in [n]}$. 
    % \mf{users report their loss with respect to one service only, right? say more clearly, emphasize such quantity can be very naturally obtained in practice (leading into the next sentence)}. 
    In practice, this step requires deploying the services and collecting signals of engagement and utility to determine the loss associated with each user, under the behavioral assumption that users are rational agents that choose the best available service. The provider thus needs to measure each user's loss \textbf{only} in their single chosen service.
    \item \textbf{User selection (line 6) } A new user $l$ is selected with probability proportional to $\mathcal{L}_i(\Theta_{t-1}, \phi_l)$. This ensures that users that are currently poorly served by existing services are more likely to be selected.
    \item \textbf{New Service (line 7-8).} Given a selected user $l$, the algorithm queries the preference $\phi_l$ of the user and centers the new service at that preference $\theta_t = \phi_l$. In practice, this step requires acquiring data about the user in order to learn their preference parameter (this is needed for only $k$ total users throughout the algorithm). For example, data may be acquired by incentivizing the selected users, via offering discount coupons or free premium subscriptions \citep{hirnschall2018learning}.
\end{itemize}

With each iteration the loss of each user is non-increasing; the previous services remained fixed and a user would switch to a new service only if it improves quality or equivalently decreases loss. Since at each iteration, a new service is added, the process terminates after $k$ steps.
Since it is costly to offer and maintain too many different services, we typically have $k \ll n$.

We now discuss the theoretical properties of the set of services we get at the termination of Algorithm~\ref{alg:initialize}.

\begin{restatable}{theorem}{main}
\label{thm:main}
    Consider $n$ users with unknown preferences $\{\phi_1, \ldots, \phi_n\} \subset \mathbb{R}^d$, and associated loss functions  $\mathcal{L}_i(\cdot,\cdot)$ satisfying Assumptions \ref{ass:unique} and \ref{ass:triangle}, with bandit access. Let $\Theta_{\rm OPT} \subset \mathbb{R}^d$ be the set of $k$ services minimizing the total loss and $\mathcal{C}(\Theta_{\rm OPT})$ the resulting partitioning of users (Definition~\ref{def:clustering}). If Algorithm \ref{alg:initialize} is used to obtain $k$ services $\Theta_k$, then the following bound holds:
    \begin{align*}
        \mathbb{E}_{\Theta_k}[\mathcal{L}(\Theta_k, [n])] &\leq  K_{\rm OPT}(2 + \log k) \cdot \mathcal{L}(\Theta_{\rm OPT}, [n]),
    \end{align*}
    where the expectation is taken over the randomization of the algorithm and $K_{\rm OPT}$ is equal to 
    \begin{equation}\label{eqn:K}
       % \max_{\mathcal{B} \in \mathcal{C}(\Theta_{\rm OPT})}
       % \frac{4}{\min\limits_{j \in \mathcal{B}}\frac{1}{|\mathcal{B}|}\sum \limits_{i \in \mathcal{B}} c_{ij}}\left(\max_{j \in \mathcal{B}} \frac{1}{|\mathcal{B}|}\sum_{i \in \mathcal{B}}\frac{1}{c_{ij}}\right).
       \Scale[1.0]{\max_{\mathcal{B} \in \mathcal{C}(\Theta_{\rm OPT})}
       \frac{4}{\min\limits_{j \in \mathcal{B}}\sum \limits_{i \in \mathcal{B}} c_{ij}}\left(\max_{j \in \mathcal{B}} \sum_{i \in \mathcal{B}}\frac{1}{c_{ij}}\right).}
    \end{equation}
\end{restatable}
A detailed proof is presented in Appendix~\ref{sec:thmmainproof}; we summarize the main ideas here. The intuition is that a chosen user's preference parameter is typically a good representative for other users in its cluster. Thus adding a service parameterized by the chosen user's preferences generally reduces the losses of users in this cluster. 
Subsequently we are less likely to pick another user from the same cluster. The $\log k$ factor is due to clusters from which users were never picked. 

A similar proof approach was used by \citet{arthur2007k} in the context of the $k$-means problem, by sequentially placing centers on \textit{known} points sampled with probability proportional to the point's squared distance to its closest existing center.
A key novelty of our analysis is to
capture the \emph{alignment} of diverse loss geometries across users in a large class of functions, specifically understanding how user similarities $c_{ij}$ affect the approximation ratio.\footnote{In addition, as stated earlier, we tackle the lack of prior information on  $\phi_i$ and the function form of $\mathcal{L}_i$---this challenge is particular to our setting and does not arise in related standard problems of clustering ($k$-means, $k$-mediods) or resource allocation (facility location problems).}
% \mf{added this to reiterate why our setting is distinct while the proof structure looks like kmeans++. it can be moved or discussed elsewhere too. Perhaps saying it in more than one place is good.}}

\textbf{Key characteristics of $K_{\rm OPT}$:} The following are essential characteristics of the term $K_{\rm OPT}$.
\begin{enumerate}[label={{(\roman*)}},%itemsep=-5pt,topsep=-5pt]
]
    \item All terms in $K_{\rm OPT}$ depend on the local clusters in the unknown optimal clustering $\mathcal{C}(\Theta_{\rm OPT})$.
    \item The constant $\min_{j \in \mathcal{B}}\frac{1}{|\mathcal{B}|}\sum_{i \in \mathcal{B}} c_{ij}$ captures the user whose loss geometry is \emph{least similar} to the average loss geometry of the cluster they belong to (recall Assumption~\ref{ass:triangle}.i).
    \item The constant $\max_{j \in \mathcal{B}} \frac{1}{|\mathcal{B}|}\sum_{i}\frac{1}{c_{ij}}$ captures the user whose preference is \emph{least similar} to the optimal service parameter of the cluster they belong to (recall Assumption~\ref{ass:triangle}.ii).
    \item Even within a cluster all terms are averages, so a few poorly aligned pairs of users don't hurt the bound if the cluster sizes are large.
\end{enumerate}

%\mf{So maybe here you can give the example of Mahalanobis distances, to give an interpretation of $K_{\rm OPT}$ for it (in terms of alignment if the $\Sigma_i$ matrices). It can a short comment, or can even be given in the appendix.}

\textbf{Fair objective.} While minimizing the total loss is beneficial from the provider's point of view in keeping users satisfied on average, it is undesirable in human-centric applications if the provided services result in unfavorable or harmful outcomes towards some demographic groups. %

Suppose the $n$ users come from $m$ different demographic groups ($m$ is typically small, say racial groups, gender). We denote the groups as $\mathcal{A} = \{\mathcal{A}_1, \ldots, \mathcal{A}_m\} \subset [n]$. The fairness objective is defined as the maximum average loss suffered by any group:
\begin{equation}\label{eq:fair}
    \Phi(\Theta, \mathcal{A}) = \max_{i \in [m]} \mathcal{L}(\Theta, \mathcal{A}_i) / |\mathcal{A}_i|.
\end{equation}
\cite{ghadiri2021socially} defined this objective in the context of fair $k$-means where the points and group identities are \textit{known} and gave a \textit{non-constructive} proof that if a $c-$approximate solution for $k$-means exists, it is $m\cdot c$-approximate for fair $k$-means. For the fairness objective, we slightly modify our algorithm, by simply reweighting the probability to select an user by the inverse of the size of their demographic group to result in Fair AcQUIre (Algorithm~\ref{alg:fair_initialize} in Appendix~\ref{sec:fair}).
% In our setup, where both user preferences and group identities are unknown,  the set of services initialized by Algorithm \ref{alg:initialize} enjoys guarantees on optimizing the fairness objective as stated below.

% \begin{restatable}{theorem}{fair}
% \label{thm:fair}
%     Consider $n$ users with unknown preferences $\{\phi_1, \ldots, \phi_n\} \subset \mathbb{R}^d$, and unknown associated loss functions  $\mathcal{L}_i(\cdot,\cdot)$ satisfying Assumptions \ref{ass:unique} and \ref{ass:triangle}. Suppose these users belong to $m$ unknown demographic groups $\mathcal{A} = \{\mathcal{A}_1, \ldots, \mathcal{A}_m\} \subset [n]$. Let $\Theta_{\rm fair} \subset \mathbb{R}^d$ be the set of $k$ services minimizing the fairness objective $\Phi$ given in \eqref{eq:fair}. If Algorithm \ref{alg:initialize} is used to obtain $k$ services $\Theta_k$, then the following bound holds:
%     \begin{align*}
%          \mathbb{E}_{\Theta_k}[\Phi(\Theta_k, \mathcal{A})] \leq m\cdot  K_{\rm OPT}\cdot  \frac{\max_{i \in [m]} |\mathcal{A}_i|}{\min_{i \in [m]} |\mathcal{A}_i|}\cdot (2 + \log k) \cdot \Phi(\Theta_{\rm fair}, \mathcal{A}),
%     \end{align*}
%     where the expectation is taken over the randomization of the algorithm and $K_{\rm OPT}$ is defined in  \eqref{eqn:K}.
% \end{restatable}
\begin{restatable}{theorem}{fairknowngroups}
\label{thm:fair_known_groups}
    Consider $n$ users with unknown preferences $\{\phi_1, \ldots, \phi_n\} \subset \mathbb{R}^d$, and associated loss functions  $\mathcal{L}_i(\cdot,\cdot)$ satisfying Assumptions \ref{ass:unique} and \ref{ass:triangle} with bandit access. Suppose these users belong to $m$ demographic groups $\mathcal{A} = \{\mathcal{A}_1, \ldots, \mathcal{A}_m\} \subset [n]$. Let $\Theta_{\rm fair} \subset \mathbb{R}^d$ be the set of $k$ services minimizing the fairness objective $\Phi$ given in \eqref{eq:fair}. If Algorithm \ref{alg:fair_initialize} is used to obtain $k$ services $\Theta_k$, then the following bound holds:
    \begin{align*}
         &\Scale[0.95]{\mathbb{E}_{\Theta_k}[\Phi(\Theta_k, \mathcal{A})] \leq m K_{\rm fair} (2 + \log k) \cdot \Phi(\Theta_{\rm fair}, \mathcal{A}),}
    \end{align*}
    where the expectation is taken over the randomization of the algorithm and $K_{\rm fair}$ is defined in  \eqref{eqn:K_scaled}.
 \end{restatable}

% If the group identities are known a priori to the provider, we can modify Algorithm~\ref{alg:initialize} to get rid of the factor $\frac{\max_{i \in [m]} |\mathcal{A}_i|}{\min_{i \in [m]} |\mathcal{A}_i|}$ in the approximation ratio, as discussed in Appendix~\ref{sec:fair}. 

\section{Generalization in Linear Predictors}\label{sec:linear}
In practical settings, a provider would want to design services that not only keep the subscribed users satisfied but also attract new users to subscribe by generalizing the services to users it has never interacted with before. Now instead of considering $n$ users, suppose that each $i \in [N]$ represents a subpopulation with its own (sub-Gaussian) distribution of features, and the provider can interact with finite samples $n_i$ from these distributions. A question that arises is whether we can deal with this finite-sample-from-subpopulations scenario, and how does the number of samples affect the algorithm’s output to unseen users. In this section, we answer this question for the special case of linear predictors (i.e., regression loss).

In this section we restrict ourselves to the special case of linear prediction tasks, where the goal is to accurately predict the score of a user as a linear function of their features. The score 
for a user in the $i^{\rm th}$ subpopulation with zero-mean random feature $x \in \mathbb{R}^d$ is generated as $y = \phi_i^\top x$ where both the true linear regressor $\phi_i \in \mathbb{R}^d$ and the feature covariance $\mathbb{E}_x [xx^\top] = \Sigma_i$ are unknown. Suppose a service uses a linear regressor $\theta \in \mathbb{R}^d$, to predict the score for this user as $\theta^\top x \in \mathbb{R}$. The loss for this subpopulation for this service is defined as the expected squared error between the predicted and actual scores, 
i.e., $\mathcal{L}_i(\theta, \phi_i) = \mathbb{E}_{(x, y)}[(\theta^\top x - y)^2] = \|\theta - \phi_i\|^2_{\Sigma_i}$.
\begin{assumption}{\label{ass:independent}}
For subpopulation $i$, features are independent draws from a zero-mean sub-Gaussian distribution. For a random feature $x \in \mathbb{R}^d$ and for any $u \in \mathbb{R}^d$, such that $\|u\|_{\Sigma_i} = 1$,
$u^\top x \in \mathbb{R}$ is sub-Gaussian with variance proxy $\sigma_i^2$.\footnote{A random variable $x$ is sub-Gaussian with variance proxy $\sigma^2$ if $\mathbb{E}[\exp(\lambda x)] \leq \exp(\frac{\sigma^2 \lambda^2}{2}) \; \forall \lambda \in \mathbb{R}$.} 
\end{assumption}
\begin{assumption}\label{ass:subpoplevelchoice}
We assume that the decision to %
choose between different services happens at a subpopulation level and not an individual level.     
\end{assumption}
To illustrate Assumption 4, consider the example of a provider that offers personalized services to schools ({subpopulations}) such as online library resources wherein the service provider queries students about the experience. Each school typically has a considerable number of students, but only a subset of them may actively respond to such queries. Once a school selects the service, it is made available to all students, and 
the provider could implement a system where students are encouraged to %
fill out a feedback form after using their service.

Suppose only $n_i$ users from subpopulation $i$ are subscribed to the services. Thus, upon choosing a service parameterized by $\theta \in \mathbb{R}^d$ the provider observes an \textit{empirical loss}, which is given by \[\Scale[0.95]{\widehat{\mathcal{L}}_i(\theta, \phi_i) = \frac{1}{n_i}\sum_{j \in [n_i]}(\theta^\top x_i^j - y_i^j)^2,}\] where  $\{(x_i^j, y_i^j)\}_{j \in [n_i]}$ are private unknown features and scores of the users. We stress that the service gets to see the value of the user loss function at the deployed $\theta$ (bandit feedback), but not the features of each subpopulation.
\begin{assumption}\label{ass:fullrank}
    The number of users from each subpopulation is greater than the dimension of the linear predictor, i.e. $n_i \geq d$ for all $i \in [N]$.
\end{assumption}

In this setting, given a set of services parameterized by $\Theta_{t-1}=\{\theta_1, \ldots, \theta_{t-1}\}$ the Steps 5-6 of Algorithm~\ref{alg:initialize} proceed with these finite sample averages $\widehat{\mathcal{L}}_i(\Theta_{t-1}, \phi_i) = \min_{j \in [t-1]}\widehat{\mathcal{L}}_i(\theta_j, \phi_i)$. In Step 7, multiple ways can be adopted by the provider to estimate $\phi_i$. Users can be given incentives to provide a batch of feature score pairs, or gradient free methods can be used to estimate the optimal solution to the regression problem. The generalization guarantee of Algorithm~\ref{alg:initialize} to the total \textit{expected loss} is stated below.

\begin{restatable}{theorem}{regression}
\label{thm:regression}
Suppose users belong to $N$ subpopulations satisfying Assumptions~\ref{ass:independent}, \ref{ass:subpoplevelchoice}, and \ref{ass:fullrank}. Let $\{n_i\}_{i \in [N]}$ denote the number of samples per subpopulation. Let $\Theta_k$ be the output of Algorithm~\ref{alg:initialize} using finite sample estimates $\widehat{\mathcal{L}}_i(\cdot, \phi_i)$ and $\Theta_{\rm OPT}$ be the optimal solution of the expected loss. Then, for any $\epsilon\in (0,1)$, %\mf{changed to $(0,1)$ from $[0,1)$}, 
if $n_i = \Omega(\frac{\sigma_i^4 \sqrt{N} \log{(2/\delta)}}{\epsilon^2})$ for all $i \in [N]$,
the following inequality holds with probability at least $1 - \delta$:
\begin{align*}
&\Scale[0.95]{\mathbb{E}_{\Theta_k}[\mathcal{L}(\Theta_k, [N])] \leq \frac{1 + \epsilon}{1 - \epsilon} K_{\rm OPT}(2 + \log k) \mathcal{L}(\Theta_{\rm OPT}, [N]), \; \text{where $K_{\rm OPT}$ is as defined in \eqref{eqn:K} and}} \\
&\Scale[0.95]{c_{ij} = \frac{1}{2}\min\ \left\{\lambda_{\mathrm{min}}(\widehat{\Sigma}_i, \widehat{\Sigma_j}), \frac{1}{\lambda_{\mathrm{min}}(\widehat{\Sigma}_i, \widehat{\Sigma_j})}\right\},
\; \text{with} \; \widehat{\Sigma}_i = \frac{1}{n_i}\sum_{l \in [n_i]}x_i^l(x_i^l)^\top,\; \widehat{\Sigma}_j = \frac{1}{n_j}\sum_{l \in [n_j]}x_j^l(x_j^l)^\top.}
\end{align*}
% with 
% \[\widehat{\Sigma}_i = \frac{1}{n_i}\sum_{l \in [n_i]}x_i^l(x_i^l)^\top\quad\text{and}\quad \widehat{\Sigma}_j = \frac{1}{n_j}\sum_{l \in [n_j]}x_j^l(x_j^l)^\top.\]
\end{restatable}
The proof is presented in Appendix~\ref{sec:finite_sample}; we provide a brief overview here. We apply the Chernoff bound to the difference between the \textit{empirical loss} and \textit{expected loss}. Note that $(i)$ even if the same set of services are provided, the loss minimizing service for the empirical loss may be different from the expected loss for any subpopulation, and $(ii)$ the optimal set of services for the total empirical loss and the total expected loss are different. Handling these carefully, and utilizing Theorem~\ref{thm:main} concludes the proof. 
\begin{remark}
    Note that $\widehat{\Sigma}_i$ is the empirical feature covariance of subpopulation $i$. The term $c_{ij}$ captures the alignment between two subpopulation's loss geometry, and here is equal to half of the \emph{minimum generalized eigenvalue} of the empirical feature covariances of the the respective subpopulations. This quantity %$c_{ij}$ 
    is the largest constant satisfying Assumption~\ref{ass:triangle} (cf. Appendix~\ref{sec:finite_sample}). 
\end{remark}

% \vskip -0.2in
\section{Experiments}\label{sec:experiments}
\begin{figure*}[t!]
\vskip -0.4in
     \centering
     \begin{subfigure}[b]{0.32\textwidth}
         \centering
         \includegraphics[width=\textwidth, trim={0.8cm 0.1cm 2.5cm 1.5cm}, clip]{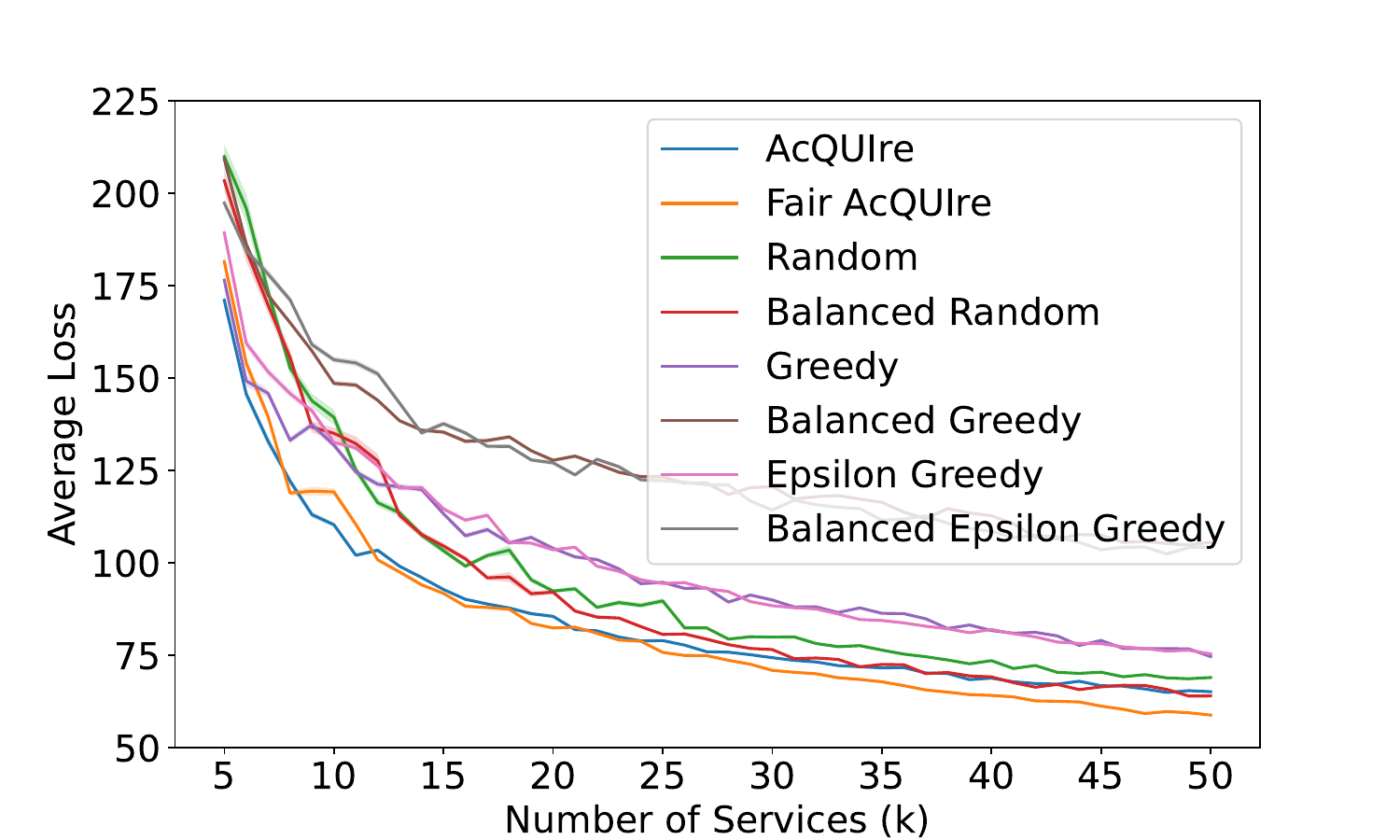}
         \caption{Avg. loss (all Census groups)}
         \label{fig:fairness_all}
     \end{subfigure}
     % \hfill
     \begin{subfigure}[b]{0.33\textwidth}
         \centering
         \includegraphics[width=\textwidth, trim={0.8cm 0.1cm 2.5cm 1.5cm}, clip]{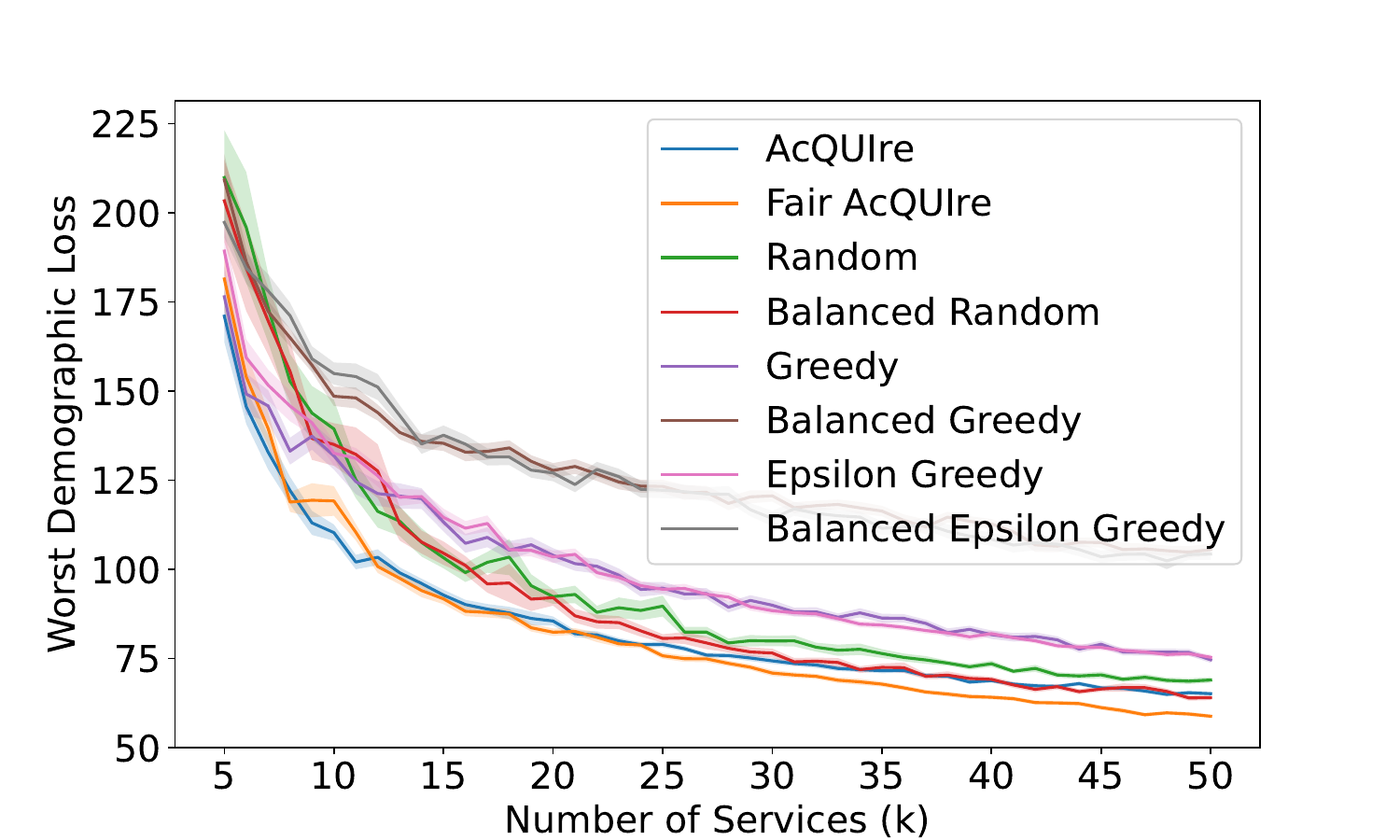}
         \caption{Avg. loss (worst demographic)}
         \label{fig:fairness_worst}
     \end{subfigure}
     % \hfill
     \begin{subfigure}[b]{0.33\textwidth}
         \centering
         \includegraphics[width=\textwidth, trim={0.8cm 0.1cm 2.5cm 1.5cm}, clip]{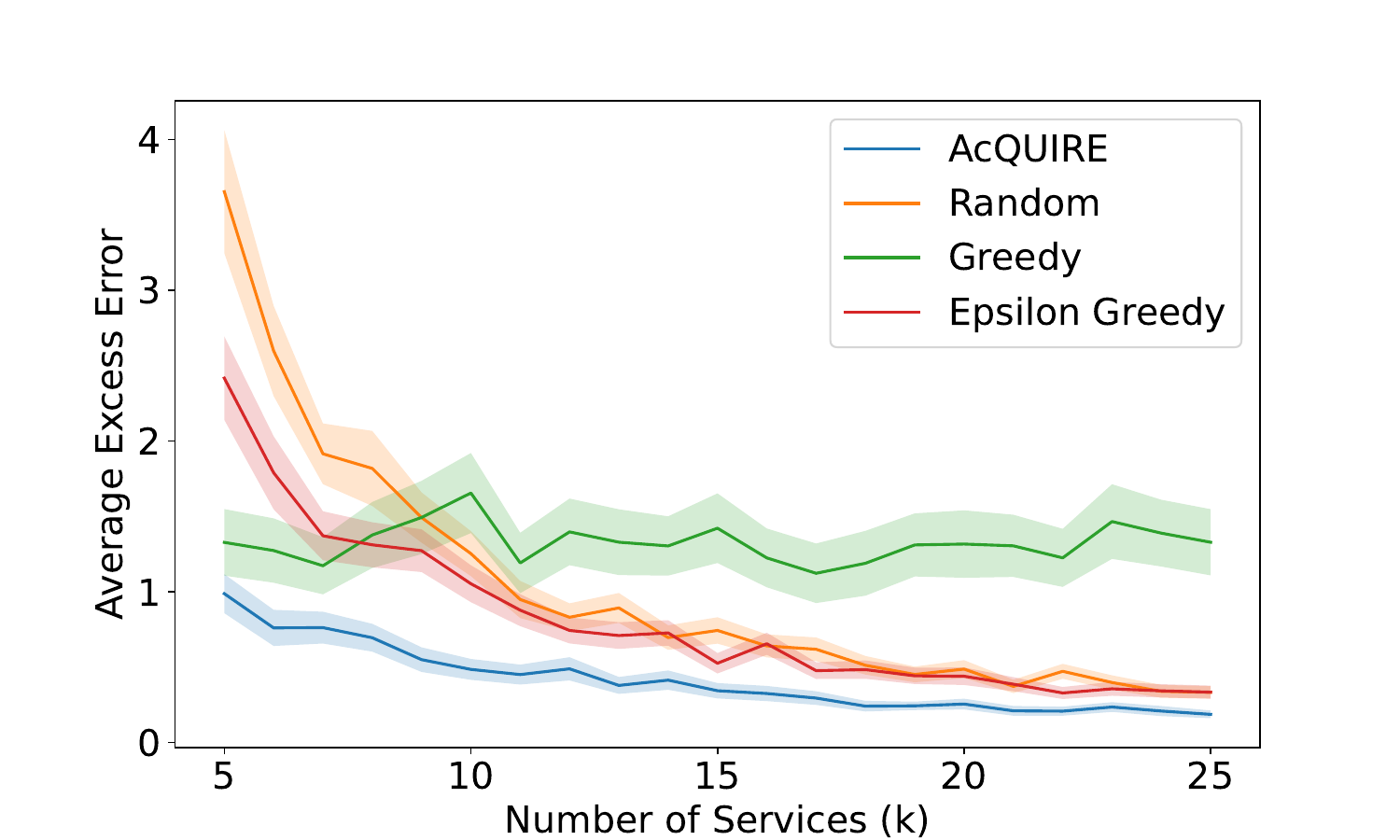}
         \caption{Avg. excess error (ML10M task)}
         \label{fig:fairness_ml10m}
     \end{subfigure}
        % \caption{(a) reports the average loss over all users while (b) reports the worst average loss across different demographics (Objective~\eqref{eq:fair}) and we observe that while the greedy and epsilon greedy baselines perform well when $k < 10$, these approaches are myopic as the random sampling strategy outperforms them when $k > 10$. AcQUIre and Fair AcQUIre and are the 2 strongest baselines for both the tasks. In (c), we report the average excess error for the movie recommendation task. We again observe that the greedy algorithm works well when $k$ is small. Epsilon Greedy opts a explore-vs-exploit approach and is able to break out of this myopicity. However, AcQUIre still emerges as the best strategy for data collection.}
        \vskip -0.1in
        \caption{
        %Fig. \ref{fig:fairness_all} presents the average loss across all users, while Fig. \ref{fig:fairness_worst} highlights the highest average loss among various demographics, as outlined in Objective~\eqref{eq:fair}. 
        Fig. \ref{fig:fairness_all} and\ref{fig:fairness_worst} show the performance of various user selection strategies on the travel time prediction task on the Census data.
        Notably, our findings reveal that the greedy and epsilon-greedy baselines exhibit strong performance for $k < 10$. However, as the value of $k$ grows, these strategies prove myopic, with random sampling surpassing their effectiveness. AcQUIre and Fair AcQUIre consistently emerge as the two best baselines for both tasks. Fig. \ref{fig:fairness_ml10m} presents the average excess error for the movie recommendation task. Remarkably, the greedy algorithm demonstrates efficacy when $k$ is small. Epsilon-greedy, employing an explore-vs-exploit approach, successfully overcomes myopic tendencies. Nevertheless, AcQUIre continues to be the best baseline for data collection.}
        \label{fig:experiments}
        \vskip -0.2in
\end{figure*}
% \begin{figure}
%   \centering
  
%   \subfigure(a){\includegraphics[width=0.24\textwidth]{monalisa.jpg}} 
%   \subfigure(b){\includegraphics[width=0.24\textwidth]{monalisa.jpg}} 
%   \subfigure(c){\includegraphics[width=0.24\textwidth]{monalisa.jpg}}
%   \subfigure(d){\includegraphics[width=0.24\textwidth]{monalisa.jpg}}

%   \caption{(a) blah (b) blah (c) blah (d) blah}
 
% \label{fig:foobar}
% \end{figure}

We empirically demonstrate\footnote{Code is available at \url{https://anonymous.4open.science/r/MultiServiceInitialization-A422}} 
% \url{https://anonymous.4open.science/r/InteractiveSystemInitialization-2A33}} 
the benefits of our algorithm on a commute time prediction task based on 2021 US Census data\footnote{\url{https://www.census.gov/programs-surveys/acs/data.html}.} and a semi-synthetic movie recommendation task on the MovieLens10M dataset. 
% In both tasks, the provider starts without access to any data and our goal is to evaluate how well the sequential acquisition of data, subsequent learning and deployment of learnt models performs on the whole dataset.
% alternative: 
In each task the multi-service provider has initially no access to data. Our goal is to evaluate the effectiveness of the iterative data collection and service initialization procedure in comparison to established baselines. Below we first describe both the tasks and then discuss the baselines we consider for our evaluation.

\textbf{Census Data.} 
 We consider the task of predicting daily work commute times, based on 2021 US census data from \textsc{folktables} \citep{ding2021retiring}. We illustrate a potential use case: the \textit{provider} is a transport authority offering \textit{services} in the form of personalized podcasts. If the duration of a service is similar to the commute time of a user, that user will be able to consume the media while travelling to work. Hence an accurate prediction of the commute time may be useful in providing services tailored to the users. 
 
\begin{figure*}[h]
     \centering
     \begin{subfigure}[b]{0.3\textwidth}
         \centering
         \includegraphics[width=\textwidth]{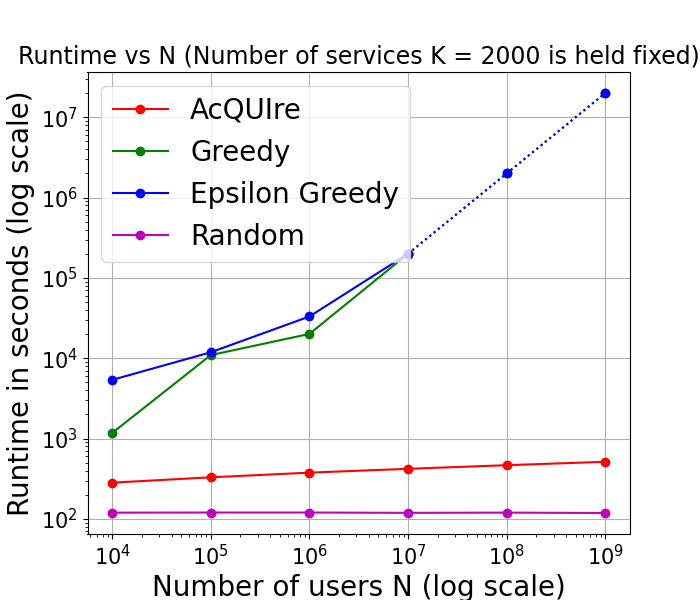}
     \end{subfigure}
     \begin{subfigure}[b]{0.3\textwidth}
         \centering
         \includegraphics[width=\textwidth]{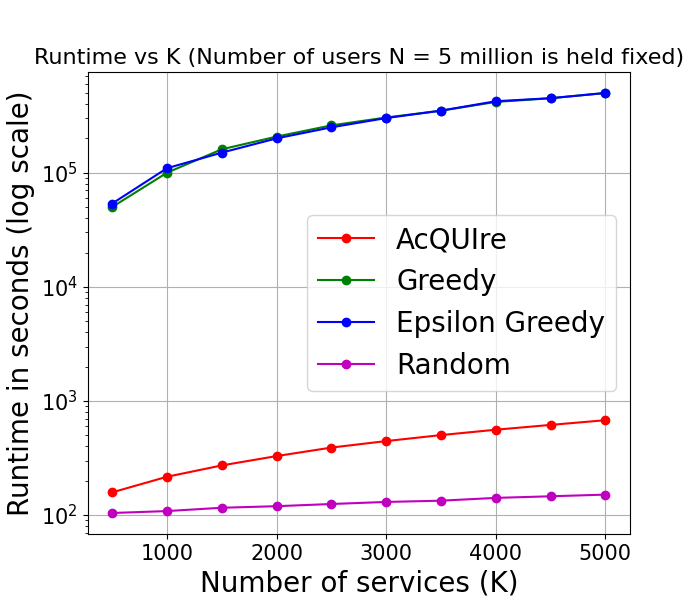}
     \end{subfigure}
        \caption{Runtimes for AcQUIre and baselines as number of users ($N$) and services ($K$) vary. 
        } 
        \label{fig:runtimes}
        % \vskip -0.2in
\end{figure*}

 The dataset has $N=7025$ subpopulations defined by area zip codes.
 We use $k$ linear predictors (as services) with user features (education, income, transportation mode, age etc). Refer to Appendix~\ref{sec:add_exp} for details on data pre-processing. The features and commute times are \textit{unknown a priori} to the provider. At time step $t \leq k$, suppose that the provider offers a set of services parameterized by $\Theta_{t-1}$. The provider observes the losses (squared prediction error of their commute times as discussed in Section \ref{sec:linear}) across different subpopulations $\{\widehat{\mathcal{L}}_i(\Theta_{t-1}, \phi_i)\}_{i \in [N]}$. Then the provider selects a subpopulation $l \in [N]$, to get the user feature and commute time data and then estimates $\phi_l$ via least squares regression. This selection can be done via our proposed method or one of the baseline strategies discussed later. The provider then centers its next service on $\phi_l$ and updates the list of offered services $\Theta_t = \Theta_{t-1} \cup \phi_l$. Note that in this process the provider only observes features and commute times of users in the $k$ selected subpopulations and $k \ll N$.
In Figure \ref{fig:runtimes} we compare runtimes, and observe that even with 1 billion users, AcQUIre %finishes running in less than 
        takes 300 sec, whereas the greedy and epsilon greedy methods take $>10^5$ sec even for 10 million users. With 5000 services, AcQUIre takes $<900$ secs, whereas the runtimes for greedy and epsilon greedy are in the range of $10^5$ secs. 
\textbf{Movie Recommendations.} We conduct a semi synthetic experiment based on the widely used Movielens10M data set \citep{harper2015movielens} containing 10000054 ratings across 10681 movies by 71567 viewers. We hold out the top $m=200$ movies and pre-process the data set to divide viewers into $N=1000$ user subpopulations based on similarity of their ratings on the remaining movies (cf. Appendix~\ref{sec:add_exp}). Our goal is to evaluate the generalization performance of the baselines to viewers that the provider has never interacted with before. Thus during the service initialization phase, only half of the users in each of the $N$ subpopulations interact with the provider (train set), and we evaluate the performance of the algorithm by the loss incurred by the initialized services on data of the other half of the users that no prior interaction with the provider (test set).  
% We compare the performance of Algorithm \ref{alg:initialize}---which initializes a set of recommendation models where user preferences are \textbf{unknown a priori}---to a uniform initialization baseline. 
% For subpopulation $i\in[N]$, let the solution to the following optimization problem denote the unknown user and item embeddings: \[(U_i, V_i) = \argmin_{(U,V) \in \mathbb{R}^{n_i \times d}\times \mathbb{R}^{d \times m}} \|(UV)_{\Omega_i} - r_i\|_2^2,\]
% where $r_i$ is the true user ratings  and $\Omega_i$ is the list of movies rated by the user for subpopulation $i$.
% User dissatisfaction for a recommendation model, parameterized by $V \in \mathbb{R}^{d \times m}$, is captured by the \textit{excess error}, namely
% \begin{equation}\label{eq:excesserror}
% \mathcal{L}_i(V, V_i) = \|(U_iV)_{\Omega_i} - r_i\|_2^2 - \|(U_iV_i)_{\Omega_i} - r_i\|_2^2.
% \end{equation}
For subpopulation $i\in[N]$, the solution to the following optimization problem 
% denote the \textit{unknown} 
denotes the user and item embedding: 
\begin{align}\label{eq:matrix_factorization}
    \Scale[0.9]{(U_i, \phi_i) = \argmin_{(U,\phi) \in \mathbb{R}^{n_i \times d}\times \mathbb{R}^{d \times m}} \|(U\phi - r_i)_{\Omega_i^{\rm train}}\|_2^2,}
\end{align}
where $r_i$ is the true user ratings  and $\Omega_i^{\rm train}$ is the list of movies rated by the users in subpopulation $i$. Since $r_i$ is a sparse matrix we consider the prediction error only on the movies they rated, i.e. $\Omega_i^{\rm train}$. 
% $(U\phi)_{\Omega_i}$ denotes those entries of the matrix $U\phi$ corresponding to $\Omega_i$. 
User loss / dissatisfaction for a recommendation model, parameterized by $\theta \in \mathbb{R}^{d \times m}$, is captured by the {excess error}, namely
\begin{equation}\label{eq:excesserror}
\Scale[0.9]{\mathcal{L}_i(\theta, \phi_i) = \|(U_i\theta - r_i)_{\Omega_i^{\rm train}}\|_2^2 - \|(U_i\phi_i - r_i)_{\Omega_i^{\rm train}}\|_2^2.}
\end{equation}
This value typically indicates how unhappy users are with the suggested movies with respect to their preferred movies. 
The provider initially doesn't know the user ratings. At time step $t \leq k$, suppose the provider offers a set of recommendation models $\Theta_{t-1}$. Users choose the service with the best recommendations and the provider observes the losses across different subpopulations $\{\mathcal{L}_i(\Theta_{t-1}, \phi_i)\}_{i \in [N]}$. Then the provider selects a subpopulation $l \in [N]$ to estimate $\phi_l$ via \eqref{eq:matrix_factorization}. This selection can be done via our proposed method or one of the baseline strategies discussed below. The provider then centers its next model on $\phi_l$ and updates the list of offered models $\Theta_t = \Theta_{t-1} \cup \phi_l$. In this process the provider only observes movie ratings of the users in the $k$ selected subpopulations. Once the services are initialized we evaluate the performance on the movies rated in the test set denoted by $\{\Omega_1^{\rm test}, \ldots, \Omega_N^{\rm test}\}$.
% The results upon using different baselines to collect data are summarized in Figure~\ref{fig:experiments}.

% Comment on baselines: a reviewer might object why are baselines only changing step 6. It has better optics if you first propose the baselines and then comment that in fact they only require 6 to change.
\textbf{Baselines.} Both our tasks iterate through the steps of observing \textit{User Behavior}, \textit{User Selection} to gather data, designing \textit{New Service} to update set of offered services. Through our experiments we wish to empirically evaluate different \textbf{User Selection} strategies with respect to AcQUIre (line 6 in Algorithm~\ref{alg:initialize}). The different user selection strategies result in the following baselines: (i) Random: $P(l = i) = 1/n$, (ii) Greedy: $l = \amax_{i \in [n] }\mathcal{L}_i(\Theta_{t-1}, \phi_i)$, (iii) Epsilon Greedy: $l = \amax_{i \in [n] }(\mathcal{L}_i(\Theta_{t-1}, \phi_i) + \epsilon_i)$ where $\epsilon_1, \ldots, \epsilon_n$ denotes zero mean i.i.d. noise. 
% As the Census Dataset has demographic groups of varying sizes (with the smallest group being 10 times smaller than the largest group), and we are also interested in the fairness objective \eqref{eq:fair} we consider incorporating these size imbalances into the algorithms. This leads to 4 additional baselines, where the choosing criteria is scaled by the corresponding group size: (v) Fair AcQUIre, (vi) Balanced Random, (vii) Balanced Greedy, (viii) Balanced Epsilon Greedy. 
% Given that census data was not introduced yet this paragraph is confusing. I it might be best to start by introducing the tasks and then the baselines
Given that the Census Dataset comprises various racial demographic groups of varying sizes (with the smallest group being ten times smaller than the largest group), and considering our interest in the fairness objective \eqref{eq:fair}, we explore incorporating these size imbalances into our algorithms. Consequently, we introduce three additional baselines, wherein the selection criteria are scaled by the corresponding group sizes: (iv) Balanced Random, (v) Balanced Greedy, and (vi) Balanced Epsilon Greedy. We benchmark them against Fair AcQUIre (Algorithm~\ref{alg:fair_initialize}) which has guarantees as as stated in  Theorem~\ref{thm:fair_known_groups}.

\textbf{Evaluation:} Each algorithm is run for 500 initialization seeds, the averages are reported in Figure~\ref{fig:experiments}.

\textbf{Runtimes: } We compare the runtimes of AcQUIre with the baselines, and study the affect of the number of users $(N)$ and number of services $(K)$ in Figure~\ref{fig:runtimes}. We find the runtimes of AcQUIre to be of the similar order of magnitude of random initialization, meanwhile performing much better than random, and the much  slower greedy and epislon greedy initialization schemes.

% \textbf{Evaluation.} We run each algorithm for 500 initialization seeds. We report the averages and discuss our findings in Figure~\ref{fig:experiments}.

% Once services are initialized, further interaction with the users gives the providers a chance to refine the services to serve the users better. A general class of loss reducing dynamics was proposed in \citep{dean2022multi} and the stability of the converged solutions was studied. In Figure~\ref{fig:importance_of_initialization} we investigate the role of initialization in (i) convergence rate, and (ii) quality of converged solution under 2 types of loss reducing dynamics. We note that AcQUIre converges faster and to better solutions compared to the baselines. 

\textbf{Impact of Initialization:}  Once a set of services are initialized, with more user interactions, the provider updates the services on new data to improve the quality (indicated by the reduction in total loss). To evaluate the importance of initialization, we conducted experiments using two different optimization algorithms: (i) Generalized k-means: The services are iteratively updated by training each service on the current group of subpopulations selecting it. After updating the service parameters, the subpopulations reselect their best service. This process repeats until convergence. (ii) Multiplicative weights update \citep{dean2022multi}: Similar to k-means, but each subpopulation can have users choosing different services simultaneously. Both generalized k-means and the multiplicative weights update guarantee that the total loss reduces over time \citep{dean2022multi}.

\begin{figure*}[h]
     \centering
     \begin{subfigure}[b]{0.245\textwidth}
         \centering
     \includegraphics[width=\textwidth , trim={0.8cm 0.1cm 3.1cm 1.5cm}, clip]{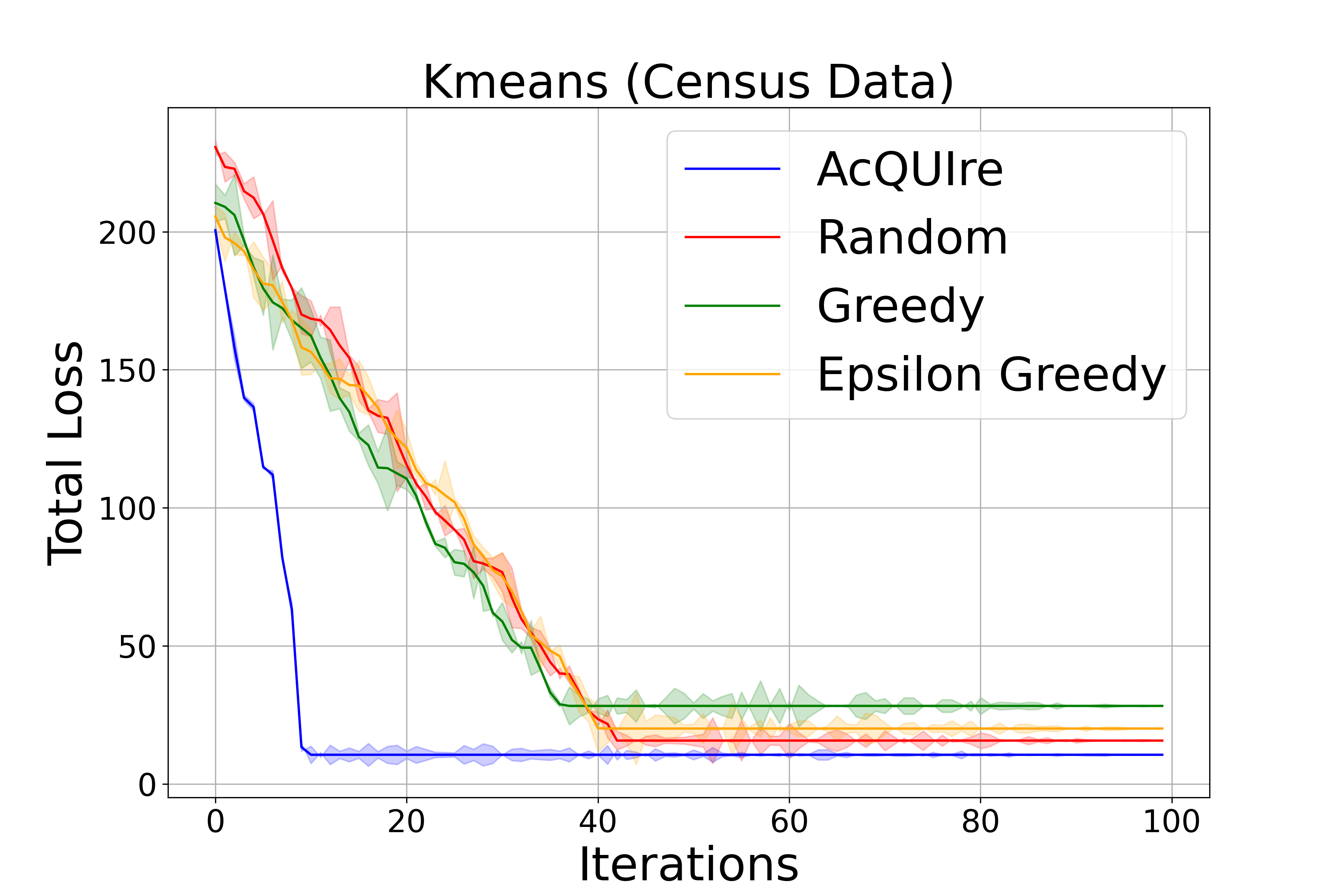}
     \end{subfigure}
     \begin{subfigure}[b]{0.245\textwidth}
         \centering
    \includegraphics[width=\textwidth, trim={0.8cm 0.1cm 3.1cm 1.5cm}, clip]{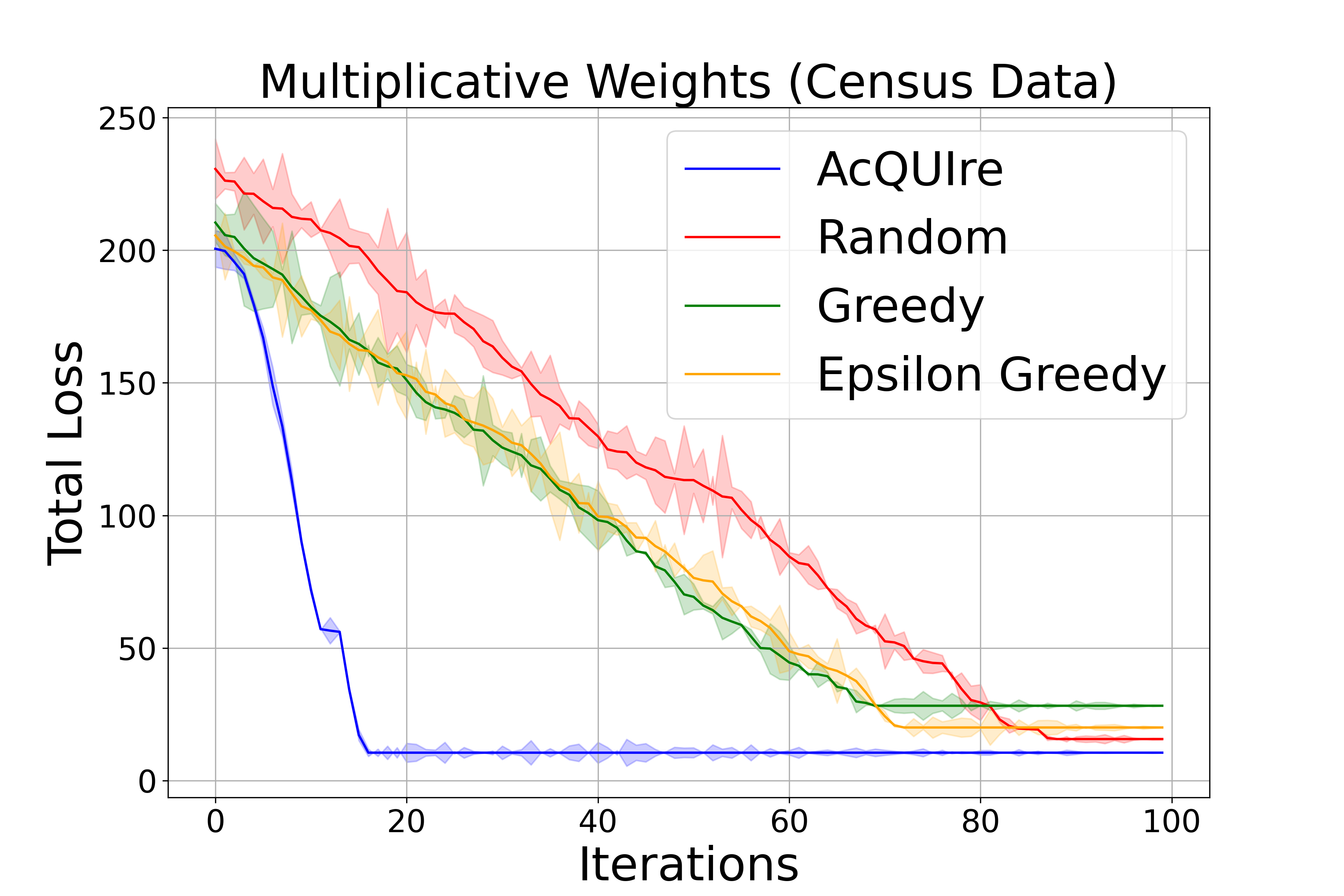}
     \end{subfigure}
     \begin{subfigure}[b]{0.245\textwidth}
         \centering         \includegraphics[width=\textwidth, trim={0.8cm 0.1cm 3.1cm 1.5cm}, clip]{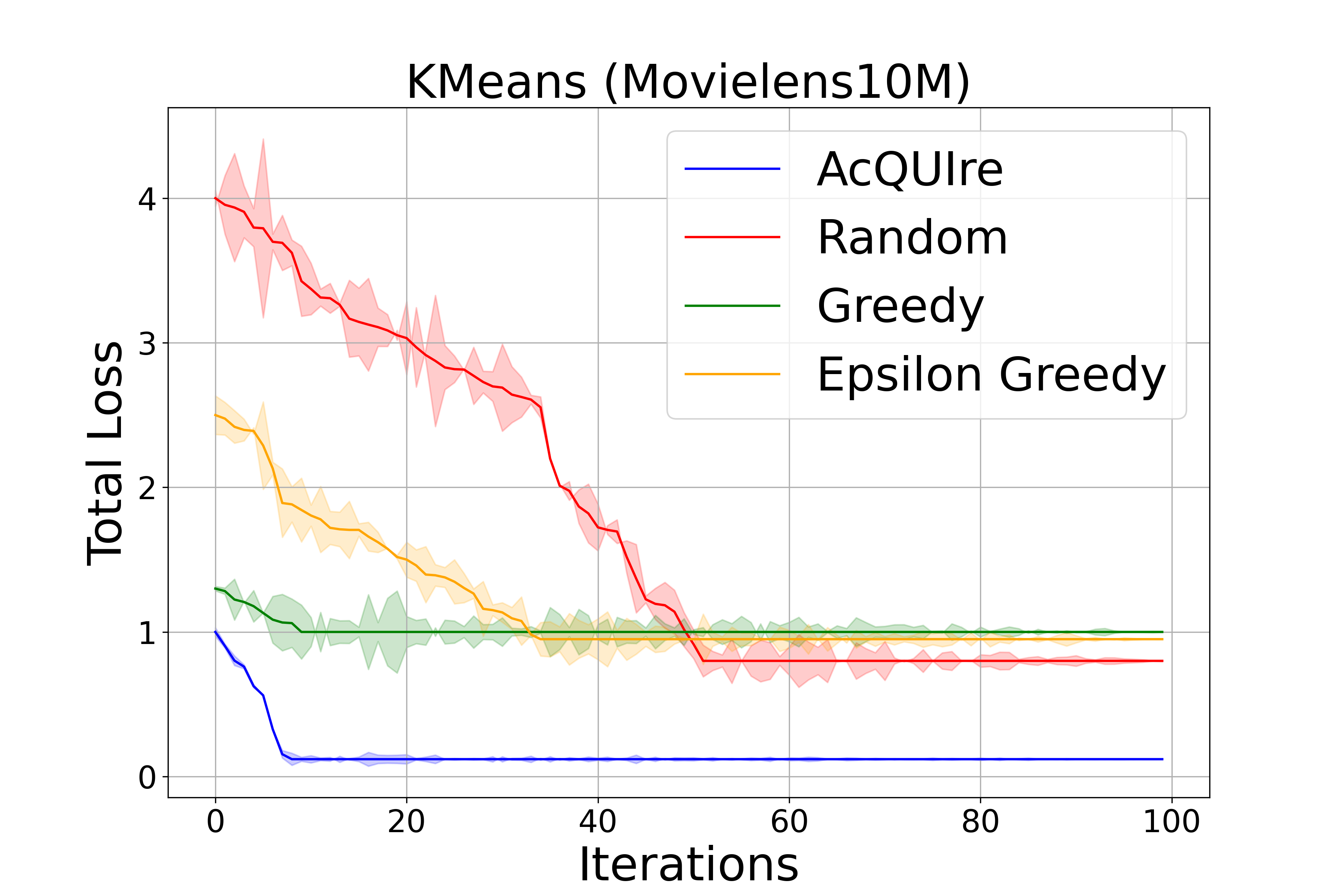}
     \end{subfigure}
     \begin{subfigure}[b]{0.245\textwidth}
         \centering        \includegraphics[width=\textwidth, trim={0.8cm 0.1cm 3.1cm 1.5cm}, clip]{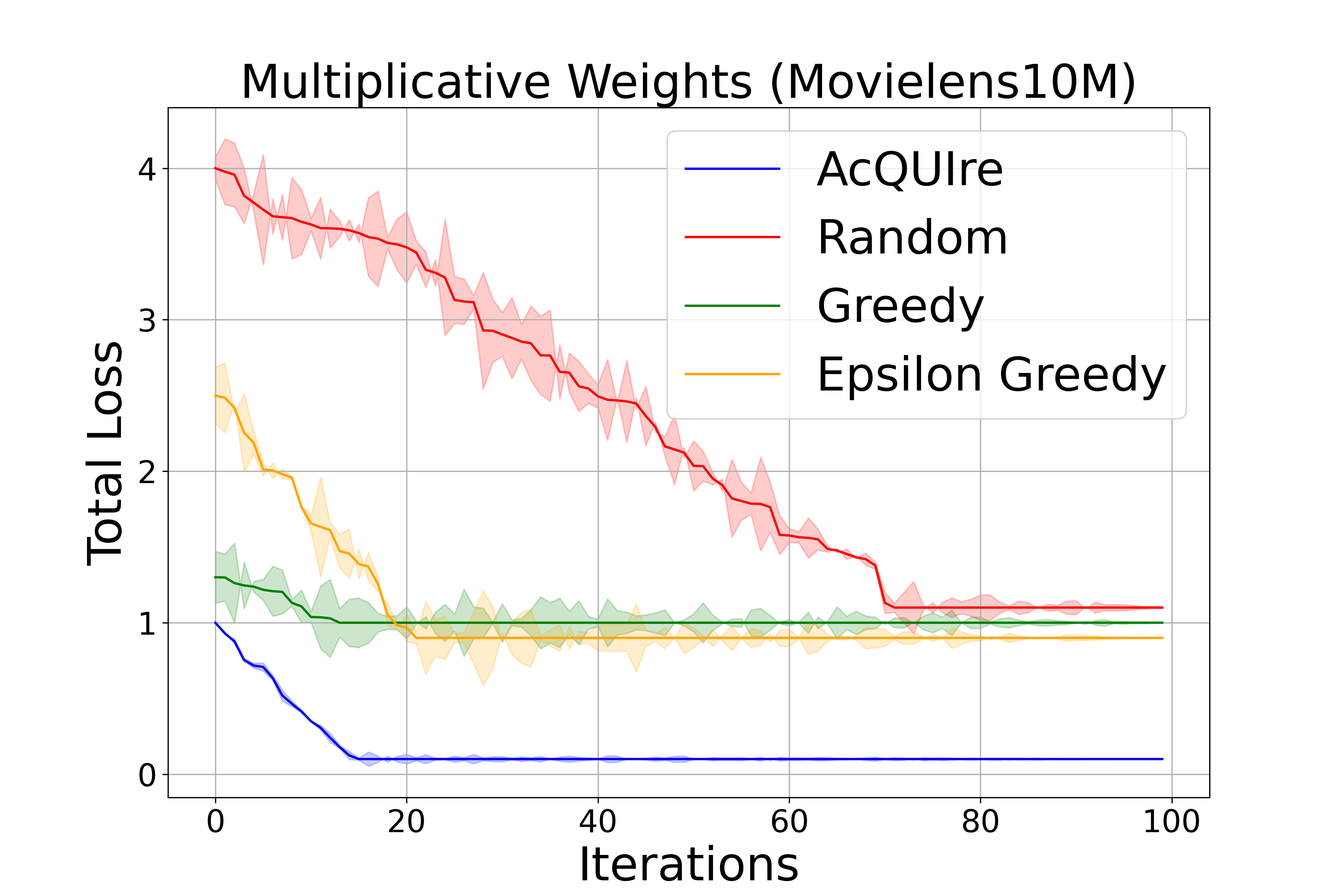}
     \end{subfigure}
        \caption{ We study the importance of initialization in both the convergence rate and quality of converged solution of optimization algorithms. We find AcQUIre converges both faster and to a lower total loss across optimization methods (kmeans and multiplicative weights) as well as datasets.}   \label{fig:importance_of_initialization}
        \vskip -0.25in
\end{figure*}

In our experiments, we initialize a set of services using AcQUIre and other baseline methods, then let both optimization algorithms run until convergence. We plot the total loss  versus the number of iterations (Figure~\ref{fig:importance_of_initialization}). Our results demonstrate that AcQUIre leads to: (1) faster convergence,
%: The optimization algorithms converge more quickly with our initialization method compared to other baselines; 
and (2) lower final loss (initializing with AcQUIre converges to lower losses; other initialization schemes are prone to being stuck in suboptimal local minima). These findings highlight the significance of a robust initialization strategy. By starting with a better initial configuration, the optimization algorithms can more effectively reach higher quality solutions.
% We chose $d=5$ for the experiment. As described for the census experiment, Algorithm~\ref{alg:initialize} iteratively selects subpopulations with probability proportional to their current excess error, and then solves the matrix factorization problem for the selected subpopulation, and adds the learnt item embedding to the list of services. 
% As shown in Figure~\ref{fig:rec}, the {total excess error} achieved by Algorithm~\ref{alg:initialize} is significantly lower than uniformly sampling users for their preferences. It is interesting to note that to achieve the same total excess error for five services initialized by our algorithm, uniform data collection and initialization would need 30 services.
\section{Conclusion}
We study the problem of initializing services for a provider catering to a user base with diverse preferences. We address the challenges of unknown user preferences, only bandit (zeroth-order) feedback from the losses, and the non-convexity of the optimization problem, by proposing an algorithm that designs services by adaptively querying data from a small set of users. We also consider the fairness aspect of such design in human centric applications. Our proposed algorithm has theoretical guarantees on both the average and fair loss objectives. 
%There are open questions relating 
Directions for future work include quantifying the robustness of the proposed initialization algorithm to noisy observations, perturbations, or outliers in the finite sample case when the feature distribution is heavy-tailed.
%which is a direction for future work. 

\section*{Acknowledgements}
Dean, Fazel, Morgenstern and Ratliff were supported in part by NSF CCF-AF 2312774/2312775. Additionally, Morgenstern's work was supported by NSF CCF 2045402. Dean's research was supported by a gift from Wayfair, a LinkedIn research award, and NSF OAC 2311521. Ratliff's research was supported by NSF CNS 1844729 and Office of Naval Research YIP Award N000142012571. Fazel was supported in part by awards NSF TRIPODS II 2023166, CCF 2007036, CCF 2212261. 
\clearpage
% \textbf{Impact Statement.} This paper presents work whose goal is to advance the field of Machine Learning. There are many potential societal consequences of our work, none which we feel must be specifically highlighted here.
\bibliography{references}

\begin{thebibliography}{42}
\providecommand{\natexlab}[1]{#1}
\providecommand{\url}[1]{\texttt{#1}}
\expandafter\ifx\csname urlstyle\endcsname\relax
  \providecommand{\doi}[1]{doi: #1}\else
  \providecommand{\doi}{doi: \begingroup \urlstyle{rm}\Url}\fi

\bibitem[vid(2005)]{vidal2005GPCA}
Generalized principal component analysis ({GPCA}).
\newblock \emph{{IEEE} Trans.\ on Pattern Analysis and Machine Intelligence},
  \penalty0 (12):\penalty0 1945–--1959, 2005.

\bibitem[Ahmadian et~al.(2019)Ahmadian, Norouzi-Fard, Svensson, and
  Ward]{ahmadian2019better}
Sara Ahmadian, Ashkan Norouzi-Fard, Ola Svensson, and Justin Ward.
\newblock Better guarantees for k-means and euclidean k-median by primal-dual
  algorithms.
\newblock \emph{SIAM Journal on Computing}, 49\penalty0 (4):\penalty0
  FOCS17--97, 2019.

\bibitem[Aloise et~al.(2009)Aloise, Deshpande, Hansen, and Popat]{aloise2009np}
Daniel Aloise, Amit Deshpande, Pierre Hansen, and Preyas Popat.
\newblock Np-hardness of euclidean sum-of-squares clustering.
\newblock \emph{Machine learning}, 75\penalty0 (2):\penalty0 245--248, 2009.

\bibitem[Arora et~al.(2016)Arora, Varshney, et~al.]{arora2016analysis}
Preeti Arora, Shipra Varshney, et~al.
\newblock Analysis of k-means and k-medoids algorithm for big data.
\newblock \emph{Procedia Computer Science}, 78:\penalty0 507--512, 2016.

\bibitem[Arthur and Vassilvitskii(2007)]{arthur2007k}
David Arthur and Sergei Vassilvitskii.
\newblock K-means++ the advantages of careful seeding.
\newblock In \emph{Proceedings of the eighteenth annual ACM-SIAM symposium on
  Discrete algorithms}, pages 1027--1035, 2007.

\bibitem[Bose et~al.(2022)Bose, Sinha, and Mai]{bose2022scalable}
Avinandan Bose, Arunesh Sinha, and Tien Mai.
\newblock Scalable distributional robustness in a class of non-convex
  optimization with guarantees.
\newblock \emph{Advances in Neural Information Processing Systems},
  35:\penalty0 13826--13837, 2022.

\bibitem[Bose et~al.(2024)Bose, Du, and Fazel]{bose2024offline}
Avinandan Bose, Simon~Shaolei Du, and Maryam Fazel.
\newblock Offline multi-task transfer rl with representational penalization.
\newblock \emph{arXiv preprint arXiv:2402.12570}, 2024.

\bibitem[Canal et~al.(2022)Canal, Mason, Vinayak, and Nowak]{canal2022one}
Gregory Canal, Blake Mason, Ramya~Korlakai Vinayak, and Robert Nowak.
\newblock One for all: Simultaneous metric and preference learning over
  multiple users.
\newblock \emph{arXiv e-prints}, pages arXiv--2207, 2022.

\bibitem[Chua et~al.(2021)Chua, Lei, and Lee]{Chua2021finetuning}
Kurtland Chua, Qi~Lei, and Jason~D. Lee.
\newblock How fine-tuning allows for effective meta-learning.
\newblock In \emph{Proc.\ of Advances in Neural Information Processing
  Systems}, volume~34, 2021.

\bibitem[Cornu{\'e}jols et~al.(1983)Cornu{\'e}jols, Nemhauser, and
  Wolsey]{cornuejols1983uncapicitated}
G{\'e}rard Cornu{\'e}jols, George Nemhauser, and Laurence Wolsey.
\newblock The uncapicitated facility location problem.
\newblock Technical report, Cornell University Operations Research and
  Industrial Engineering, 1983.

\bibitem[Dasgupta(2008)]{dasgupta2008hardness}
Sanjoy Dasgupta.
\newblock \emph{The hardness of k-means clustering}.
\newblock Department of Computer Science and Engineering, University of
  California, San Diego, 2008.

\bibitem[Dean et~al.(2022)Dean, Curmei, Ratliff, Morgenstern, and
  Fazel]{dean2022multi}
Sarah Dean, Mihaela Curmei, Lillian~J Ratliff, Jamie Morgenstern, and Maryam
  Fazel.
\newblock Multi-learner risk reduction under endogenous participation dynamics.
\newblock \emph{arXiv preprint arXiv:2206.02667}, 2022.

\bibitem[Ding et~al.(2021)Ding, Hardt, Miller, and Schmidt]{ding2021retiring}
Frances Ding, Moritz Hardt, John Miller, and Ludwig Schmidt.
\newblock Retiring adult: New datasets for fair machine learning.
\newblock \emph{Advances in Neural Information Processing Systems},
  34:\penalty0 6478--6490, 2021.

\bibitem[F{\"o}rstner and Moonen(2003)]{forstner2003metric}
Wolfgang F{\"o}rstner and Boudewijn Moonen.
\newblock A metric for covariance matrices.
\newblock \emph{Geodesy-the Challenge of the 3rd Millennium}, pages 299--309,
  2003.

\bibitem[Ghadiri et~al.(2021)Ghadiri, Samadi, and Vempala]{ghadiri2021socially}
Mehrdad Ghadiri, Samira Samadi, and Santosh Vempala.
\newblock Socially fair k-means clustering.
\newblock In \emph{Proceedings of the 2021 ACM Conference on Fairness,
  Accountability, and Transparency}, pages 438--448, 2021.

\bibitem[Ghojogh et~al.(2019)Ghojogh, Karray, and
  Crowley]{ghojogh2019eigenvalue}
Benyamin Ghojogh, Fakhri Karray, and Mark Crowley.
\newblock Eigenvalue and generalized eigenvalue problems: Tutorial.
\newblock \emph{arXiv preprint arXiv:1903.11240}, 2019.

\bibitem[Ghosh et~al.(2020)Ghosh, Chung, Yin, and
  Ramchandran]{ghosh2020efficient}
Avishek Ghosh, Jichan Chung, Dong Yin, and Kannan Ramchandran.
\newblock An efficient framework for clustered federated learning.
\newblock \emph{Advances in Neural Information Processing Systems},
  33:\penalty0 19586--19597, 2020.

\bibitem[Ginart et~al.(2021)Ginart, Zhang, Kwon, and Zou]{ginart2021competing}
Tony Ginart, Eva Zhang, Yongchan Kwon, and James Zou.
\newblock Competing ai: How does competition feedback affect machine learning?
\newblock In \emph{International Conference on Artificial Intelligence and
  Statistics}, pages 1693--1701. PMLR, 2021.

\bibitem[Harper and Konstan(2015)]{harper2015movielens}
F~Maxwell Harper and Joseph~A Konstan.
\newblock The movielens datasets: History and context.
\newblock \emph{Acm transactions on interactive intelligent systems (tiis)},
  5\penalty0 (4):\penalty0 1--19, 2015.

\bibitem[Hashimoto et~al.(2018)Hashimoto, Srivastava, Namkoong, and
  Liang]{hashimoto2018fairness}
Tatsunori Hashimoto, Megha Srivastava, Hongseok Namkoong, and Percy Liang.
\newblock Fairness without demographics in repeated loss minimization.
\newblock In \emph{International Conference on Machine Learning}, pages
  1929--1938. PMLR, 2018.

\bibitem[Hirnschall et~al.(2018)Hirnschall, Singla, Tschiatschek, and
  Krause]{hirnschall2018learning}
Christoph Hirnschall, Adish Singla, Sebastian Tschiatschek, and Andreas Krause.
\newblock Learning user preferences to incentivize exploration in the sharing
  economy.
\newblock In \emph{Proceedings of the AAAI Conference on Artificial
  Intelligence}, volume~32, 2018.

\bibitem[Honorio and Jaakkola(2014)]{honorio2014tight}
Jean Honorio and Tommi Jaakkola.
\newblock Tight bounds for the expected risk of linear classifiers and
  pac-bayes finite-sample guarantees.
\newblock In \emph{Artificial Intelligence and Statistics}, pages 384--392.
  PMLR, 2014.

\bibitem[Hug(2020)]{Hug2020}
Nicolas Hug.
\newblock Surprise: A python library for recommender systems.
\newblock \emph{Journal of Open Source Software}, 5\penalty0 (52):\penalty0
  2174, 2020.
\newblock \doi{10.21105/joss.02174}.
\newblock URL \url{https://doi.org/10.21105/joss.02174}.

\bibitem[Kanungo et~al.(2002)Kanungo, Mount, Netanyahu, Piatko, Silverman, and
  Wu]{kanungo2002local}
Tapas Kanungo, David~M Mount, Nathan~S Netanyahu, Christine~D Piatko, Ruth
  Silverman, and Angela~Y Wu.
\newblock A local search approximation algorithm for k-means clustering.
\newblock In \emph{Proceedings of the eighteenth annual symposium on
  Computational geometry}, pages 10--18, 2002.

\bibitem[Kong et~al.(2020)Kong, Somani, Song, Kakade, and Oh]{kong2020meta}
Weihao Kong, Raghav Somani, Zhao Song, Sham Kakade, and Sewoong Oh.
\newblock Meta-learning for mixed linear regression.
\newblock In \emph{International Conference on Machine Learning}, pages
  5394--5404. PMLR, 2020.

\bibitem[Lattanzi and Sohler(2019)]{lattanzi2019better}
Silvio Lattanzi and Christian Sohler.
\newblock A better k-means++ algorithm via local search.
\newblock In \emph{International Conference on Machine Learning}, pages
  3662--3671. PMLR, 2019.

\bibitem[Li et~al.(2020)Li, Sahu, Talwalkar, and Smith]{li2020federated}
Tian Li, Anit~Kumar Sahu, Ameet Talwalkar, and Virginia Smith.
\newblock Federated learning: Challenges, methods, and future directions.
\newblock \emph{IEEE signal processing magazine}, 37\penalty0 (3):\penalty0
  50--60, 2020.

\bibitem[Lloyd(1982)]{lloyd1982least}
Stuart Lloyd.
\newblock Least squares quantization in pcm.
\newblock \emph{IEEE transactions on information theory}, 28\penalty0
  (2):\penalty0 129--137, 1982.

\bibitem[Makarychev et~al.(2020)Makarychev, Reddy, and
  Shan]{makarychev2020improved}
Konstantin Makarychev, Aravind Reddy, and Liren Shan.
\newblock Improved guarantees for k-means++ and k-means++ parallel.
\newblock \emph{Advances in Neural Information Processing Systems},
  33:\penalty0 16142--16152, 2020.

\bibitem[Mansour et~al.(2020)Mansour, Mohri, Ro, and Suresh]{mansour2020three}
Yishay Mansour, Mehryar Mohri, Jae Ro, and Ananda~Theertha Suresh.
\newblock Three approaches for personalization with applications to federated
  learning.
\newblock \emph{arXiv preprint arXiv:2002.10619}, 2020.

\bibitem[Masoudnia and Ebrahimpour(2014)]{masoudnia2014mixture}
Saeed Masoudnia and Reza Ebrahimpour.
\newblock Mixture of experts: a literature survey.
\newblock \emph{The Artificial Intelligence Review}, 42\penalty0 (2):\penalty0
  275, 2014.

\bibitem[Narang et~al.(2022)Narang, Sadeghi, Ratliff, Fazel, and
  Bilmes]{narang2022online}
Adhyyan Narang, Omid Sadeghi, Lillian~J Ratliff, Maryam Fazel, and Jeff Bilmes.
\newblock Online submodular+ supermodular (bp) maximization with bandit
  feedback.
\newblock \emph{arXiv preprint arXiv:2207.03091}, 2022.

\bibitem[Ostrovsky et~al.(2013)Ostrovsky, Rabani, Schulman, and
  Swamy]{ostrovsky2013effectiveness}
Rafail Ostrovsky, Yuval Rabani, Leonard~J Schulman, and Chaitanya Swamy.
\newblock The effectiveness of lloyd-type methods for the k-means problem.
\newblock \emph{Journal of the ACM (JACM)}, 59\penalty0 (6):\penalty0 1--22,
  2013.

\bibitem[Sattler et~al.(2020)Sattler, M{\"u}ller, and
  Samek]{sattler2020clustered}
Felix Sattler, Klaus-Robert M{\"u}ller, and Wojciech Samek.
\newblock Clustered federated learning: Model-agnostic distributed multitask
  optimization under privacy constraints.
\newblock \emph{IEEE transactions on neural networks and learning systems},
  32\penalty0 (8):\penalty0 3710--3722, 2020.

\bibitem[Song et~al.(2014)Song, Tekin, and Van Der~Schaar]{song2014online}
Linqi Song, Cem Tekin, and Mihaela Van Der~Schaar.
\newblock Online learning in large-scale contextual recommender systems.
\newblock \emph{IEEE Transactions on Services Computing}, 9\penalty0
  (3):\penalty0 433--445, 2014.

\bibitem[Steinhardt et~al.(2016)Steinhardt, Valiant, and
  Charikar]{steinhardt2016avoiding}
Jacob Steinhardt, Gregory Valiant, and Moses Charikar.
\newblock Avoiding imposters and delinquents: Adversarial crowdsourcing and
  peer prediction.
\newblock \emph{Advances in Neural Information Processing Systems}, 29, 2016.

\bibitem[Sun et~al.(2021)Sun, Narang, Gulluk, Oymak, and Fazel]{sun2021towards}
Yue Sun, Adhyyan Narang, Ibrahim Gulluk, Samet Oymak, and Maryam Fazel.
\newblock Towards sample-efficient overparameterized meta-learning.
\newblock In \emph{Proc.\ of Advances in Neural Information Processing
  Systems}, volume~34, 2021.

\bibitem[Tatli et~al.(2022)Tatli, Nowak, and Vinayak]{tatli2022learning}
Gokcan Tatli, Rob Nowak, and Ramya~Korlakai Vinayak.
\newblock Learning preference distributions from distance measurements.
\newblock In \emph{2022 58th Annual Allerton Conference on Communication,
  Control, and Computing (Allerton)}, pages 1--8. IEEE, 2022.

\bibitem[Ustun et~al.(2019)Ustun, Liu, and Parkes]{ustun2019fairness}
Berk Ustun, Yang Liu, and David Parkes.
\newblock Fairness without harm: Decoupled classifiers with preference
  guarantees.
\newblock In \emph{International Conference on Machine Learning}, pages
  6373--6382. PMLR, 2019.

\bibitem[Vershynin(2018)]{vershynin2018high}
Roman Vershynin.
\newblock \emph{High-dimensional probability: An introduction with applications
  in data science}, volume~47.
\newblock Cambridge university press, 2018.

\bibitem[Zhang et~al.(2019)Zhang, Khaliligarekani, Tekin, and
  Liu]{zhang2019group}
Xueru Zhang, Mohammadmahdi Khaliligarekani, Cem Tekin, and Mingyan Liu.
\newblock Group retention when using machine learning in sequential decision
  making: the interplay between user dynamics and fairness.
\newblock \emph{Advances in Neural Information Processing Systems}, 32, 2019.

\bibitem[Zhong et~al.(2016)Zhong, Jain, and Dhillon]{zhong2016mixed}
Kai Zhong, Prateek Jain, and Inderjit~S Dhillon.
\newblock Mixed linear regression with multiple components.
\newblock \emph{Advances in neural information processing systems}, 29, 2016.

\end{thebibliography}
\bibliographystyle{plainnat}

%%%%%%%%%%%%%%%%%%%%%%%%%%%%%%%%%%%%%%%%%%%%%%%%%%%%%%%%%%%%%%%%%%%%%%%%%%%%%%%
%%%%%%%%%%%%%%%%%%%%%%%%%%%%%%%%%%%%%%%%%%%%%%%%%%%%%%%%%%%%%%%%%%%%%%%%%%%%%%%
% APPENDIX
%%%%%%%%%%%%%%%%%%%%%%%%%%%%%%%%%%%%%%%%%%%%%%%%%%%%%%%%%%%%%%%%%%%%%%%%%%%%%%%
%%%%%%%%%%%%%%%%%%%%%%%%%%%%%%%%%%%%%%%%%%%%%%%%%%%%%%%%%%%%%%%%%%%%%%%%%%%%%%%
\newpage

\newpage
\appendix
\onecolumn
\section{More Discussion on Related Works}\label{sec:additional_related_works}
\textbf{Retention.} User retention in machine learning systems is closely related to the decision dynamics between the provider and users studied in the single service setting by \citet{hashimoto2018fairness, zhang2019group} and multiple services setting by \citet{dean2022multi, ginart2021competing}. In settings with multiple
sub-populations of users of different types, the question of retention has been explored in parallel
with the issue of fairness. These works typically focus on the stability of the dynamics at equilibrium. However, the final outcome is heavily influenced by the initial data the provider has and the initial configuration (partitioning) of the users across the offered services. Our work addresses \textit{the impact of initialization on the final outcome with theoretical guarantees.}

\textbf{Mixture of Experts.} This model specialization is also explored in the mixture of experts literature (see, e.g., \cite{masoudnia2014mixture}), which uses multiple `expert' models to enhance accuracy and robustness, assigning inputs to the most suitable expert based on their features through a gating mechanism.

\textbf{Clustering.} 
% and showed a family of $k$-means instances where the  approximation ratio is $2\log k$ showing their analysis is almost tight. 
\cite{makarychev2020improved} improved the analysis of \citep{arthur2007k} to show an approximation ratio $5(2 + \log k)$ and showed a family of instances with $5 \log k$ approximation ratio, thus showing $k$-means++ is tight. Recent works in clustering \citep{kanungo2002local,ahmadian2019better,lattanzi2019better} provide a constant approximation ratio, and although they are polynomial time algorithms, these methods are data inefficient and rely on \textit{knowing} the points \emph{a priori}.  
% Finally, another common clustering algorithm is Gaussian mixture models (GMMs), where each cluster is represented by a Gaussian distribution, thus clusters themselves can be viewed as ellipsoidal (which is a different goal from ours even we choose Mahalanobis distances as losses, for us in this case every point (user) is associated with a different loss geometry).

\textbf{Facility Location Problem.} Our algorithm has some resemblances with the facility location problem where there are $n$ users,  and a provider can set up at most $k$ facilities at one of $m$ candidate locations (also known as the $k$-medoids objective \citep{arora2016analysis}). One key difference is the provider can choose from $m$ predecided locations, compared to our setting where the optimization space is infinite. However our algorithm initializes only at $k$ of the $n$ user preferences, hence it can be viewed as the candidate locations simply being the user preferences and thus $m=n$. A typical greedy algorithm for the $k$-medoids objective proceeds by evaluating the marginal decrement in total loss for \emph{all possible candidates} and selects the candidate with maximal loss reduction. Thus (i) the algorithm would need to know all $\{\phi_1, \ldots, \phi_n\}$ apriori, and (ii) deploy $n$ services to obtain all $n^2$ function evaluations $\mathcal{L}_j(\phi_i, \phi_j), \;\forall i,j \in [n]$. This is infeasible in applications like online recommendation systems where $n$ is very large and typically a provider has the capacity to deploy only  $k \ll n$ services. Hence a $k$-medoids like objective will not be reasonable in our setup with incomplete information. 

\cite{bose2022scalable} studied the case where the provider had prior access to only $N < n$ users' utility functions before deploying services and the goal was to minimize the worst case total (over all $n$ users) loss. However their results assume they can solve a computationally hard mixed integer program optimally.

\textbf{Preference Learning.}
Given a \emph{fixed} set of items or services, \cite{tatli2022learning} focus on learning preference distributions of users. \cite{canal2022one} extend this to the setting where the user losses are given by an identical metric, and they learn both preferences and the metric efficiently. Our setting focuses on the \textit{design of services rather than learning preferences over a fixed set of services, and we also allow each user to have a different loss function and loss geometry}.
% ; e.g., Mahalanobis distances with different matrices (where the sublevel sets of the loss are ellipsoids with different principal axes and orientations).

\section{Motivating Example}\label{appendix:example}
% \mf{we can move this paragraph to the appendix and enhance it with the netflix example from the rebuttal as a concrete recommender system setup.} 
We conceptualize `services' broadly, encompassing both sets of independent learners—such as various service providers collaborating on initialization—and single learners with multiple models, like companies with diverse platforms or multi-model servers in federated learning settings. This initialization process is further relevant for social planners aiming to facilitate the coordination of service initialization.

\textbf{Netflix Example.} All details are borrowed from the actual working of Netflix (refer \url{https://recoai.net/netflix-recommendation-system-how-it-works/} for a detailed description and more references). The Netflix homepage displays several rows of suggestions. Each row is a collection of movies and the rows are arranged from top to bottom in decreasing order of likelihood that the user will pick movies to watch. The user has a complete choice of which row to select a movie to watch from. The different rows are generated by different underlying recommendation models parametrized by ${\theta_1, \ldots, \theta_K}$. Our model abstracts these rows out as services. Netflix maintains $K=1300$ such different recommendation models. Users' decision of watching a movie from the available rows of recommendations on the Netflix homepage informs the provider (Netflix) which of the rows (parameterized recommendation model) the user prefers most. This let's them finetune their recommendation models.
\section{Mappings satisfying Assumptions~\ref{ass:unique} and \ref{ass:triangle}}\label{sec:examples}
In this appendix, we provide more details on examples that satisfy Assumptions~\ref{ass:unique} and \ref{ass:triangle}.
For the \textbf{squared error loss for linear predictors}, we refer the reader to  Lemma~\ref{lem:geneigenvalue} for a derivation of $c_{ij}$. The remainder of the example classes are detailed below.

\subsection{Huber Loss}
Recall that the Huber loss is defined by
\begin{equation*}
    \mathcal{L}_i(\theta, \phi_i) =
    \begin{cases}
        \frac{1}{2} \|\theta - \phi_i\|^2 & \text{if } \|\theta - \phi_i\| \leq \delta, \\
         \delta(\|\theta - \phi_i\| - \frac{1}{2}\delta)& \text{otherwise}.
    \end{cases}
\end{equation*}
Note that the Huber loss varies quadratically when the error $\|\theta - \phi_i\| \leq \delta$ and linearly otherwise. We will refer to these as quadratic and linear regime of the Huber loss subsequently.

We divide the derivation into 2 subcases.

\noindent\textbf{Case 1.} $\|\theta - \phi_i\| \leq 2\delta$. Within \textbf{Case 1}, there are several subcases to consider. We analyze each one separately. 
\begin{itemize}
    \item \textbf{At least one of the terms on the right is linear:} Under the condition of \textbf{Case 1}, $\mathcal{L}_i(\theta, \phi_i) \leq \frac{3}{2}\delta^2$. Note that the Huber loss is monotonicly increasing in the prediction error.   The function value in the linear in $\|\theta - \phi_i\|$ is at least $\frac{1}{2}\delta^2$. Thus, if one of the terms in $\mathcal{L}_j(\theta, \phi_j) + \mathcal{L}_j(\phi_i, \phi_j)$ were in the linear regime, the $\mathcal{L}_j(\theta, \phi_j) + \mathcal{L}_j(\phi_i, \phi_j) \geq \frac{1}{2}\delta^2$ and $\frac{1}{3}\mathcal{L}_i(\theta, \phi_i)  \leq \mathcal{L}_j(\theta, \phi_j) + \mathcal{L}_j(\phi_i, \phi_j)$. 
    \item \textbf{Both terms on the right are quadratic:} In this sub-case, it needs to be shown when both terms of $\mathcal{L}_j(\theta, \phi_j) + \mathcal{L}_j(\phi_i, \phi_j)$ are in the quadratic regime the bound still holds.

By the triangle inequality on norms, we have that
\begin{align*}
    \|\theta - \phi_i\| \leq \|\theta - \phi_j\| + \|\phi_i - \phi_j\|.
\end{align*}
Using the power mean inequality, namely
\[a \leq b + c \implies a^2 \leq 2(b^2 + c^2),\] 
on the triangle inequality, we deduce the following implication:
\begin{align*}
    \|\theta - \phi_i\| \leq \|\theta - \phi_j\| + \|\phi_i - \phi_j\|\ \implies\  \frac{1}{2}\|\theta - \phi_i\|^2 \leq \|\theta - \phi_j\|^2 + \|\phi_i - \phi_j\|^2.
\end{align*}
If the $\mathcal{L}_i(\theta, \phi_i)$ is quadratic, we  have that $\frac{1}{2}\mathcal{L}_i(\theta, \phi_i)  \leq \mathcal{L}_j(\theta, \phi_j) + \mathcal{L}_j(\phi_i, \phi_j)$.
Now, let $\mathcal{L}_i(\theta, \phi_i)$ be linear. Then, we deduce that 
\begin{align*}
    \mathcal{L}_j(\theta, \phi_j) + \mathcal{L}_j(\phi_i, \phi_j) 
    = \frac{1}{2}(\|\theta - \phi_j\|^2 + \|\phi_i - \phi_j\|^2)
    \geq \frac{1}{4} \|\theta - \phi_i\|^2.
\end{align*}
Since the equation $\frac{1}{4}x^2 - \frac{1}{3}x + \frac{1}{6}=0$ has no real roots, the expression is always positive. Hence, by substituting $x = \frac{\|\theta - \phi_i\|}{\delta}$, we have that 
\begin{align*}
    \frac{1}{4} \|\theta - \phi_i\|^2 \geq \frac{1}{3}\delta(\|\theta - \phi_i\| - \frac{1}{2}\delta)
    = \frac{1}{3}\mathcal{L}_i(\theta, \phi_i).
\end{align*}
Therefore, we deduce that  \[\frac{1}{3}\mathcal{L}_i(\theta, \phi_i) \leq \mathcal{L}_j(\theta, \phi_j) + \mathcal{L}_j(\phi_i, \phi_j).\]
\end{itemize}

\noindent\textbf{Case 2.} $\|\theta - \phi_i\| > 2\delta$. First, observe that 
\begin{align*}
    \frac{1}{3}\mathcal{L}_i(\theta, \phi_i) &= \frac{1}{3}\delta(\|\theta - \phi_i\| - \frac{1}{2}\delta)\\
    &= \frac{1}{3} \delta \|\theta - \phi_i\| + \frac{1}{6} \delta \|\theta - \phi_i\| - \frac{1}{6} \delta \|\theta - \phi_i\| - \frac{1}{6}\delta^2\\
    &\leq \frac{1}{3} \delta \|\theta - \phi_i\| + \frac{1}{6} \delta \|\theta - \phi_i\| - \frac{1}{3} \delta^2 - \frac{1}{6}\delta^2 \\
    &\leq \frac{1}{2} \delta \|\theta - \phi_i\| - \frac{1}{2}\delta^2,
\end{align*}
where the second to last inequality follows from the fact that  $\|\theta - \phi_i\| > 2\delta$.
By the triangle inequality on norms, we have that
\begin{align*}
    \|\theta - \phi_i\| \leq \|\theta - \phi_j\| + \|\phi_i - \phi_j\|.
\end{align*} In turn, this implies either $\|\theta - \phi_j\|$ or $\|\phi_i - \phi_j\|$ is greater than $\frac{1}{2}\|\theta - \phi_i\|$.
Without loss of generality, assume $\|\theta - \phi_j\| \geq \frac{1}{2}\|\theta - \phi_i\| \geq \delta$.
Thus \[L_j(\theta, \phi_j) = \delta(\|\theta - \phi_j\| - \frac{1}{2}\delta) \geq \frac{1}{2} \delta \|\theta - \phi_i\| - \frac{1}{2}\delta^2 \geq \frac{1}{3} \mathcal{L}_i(\theta, \phi_i).\] 
Since $\mathcal{L}_j(\phi_i, \phi_j) \geq 0$, we have that
\begin{align*}
    \frac{1}{3}\mathcal{L}_i(\theta, \phi_i) \leq \mathcal{L}_j(\theta, \phi_j) + \mathcal{L}_j(\phi_i, \phi_j).
\end{align*}

\subsection{The normalized cosine distance} Recall that the normalized cosine distance is given by \[\mathcal{L}_i(\theta, \phi_i) = 1 - \theta^\top \phi_i\quad\text{where} \quad  \|\theta\|_2 = \|\phi_i\|_2 = 1.\]
Therefore, we have that
\begin{align*}
    1 - \theta^\top \phi_i=\frac{1}{2}(\|\theta\|_2^2 + \|\phi_i\|_2^2 - 2\theta^\top \phi_i)= \frac{1}{2}\|\theta - \phi\|_2^2.
\end{align*}
Assumption~\ref{ass:triangle} is satisfied with $c_{ij} = \frac{1}{2}$ by using triangle inequality followed by power mean inequality.

\subsection{Mahalanobis distance} Consider the Mahalanobis distance which is defined by \[\mathcal{L}_i(\theta, \phi_i) = \|\theta - \phi_i\|_{\Sigma_i},\] where  $\Sigma_i$ is full rank and $c_{ij} = \min\{\lambda_{\rm min}(\Sigma_i, \Sigma_j), \frac{1}{\lambda_{\rm min}(\Sigma_i, \Sigma_j)}\}$.
The derivation is similar to proof of Lemma~\ref{lem:geneigenvalue}.

\section{Proof of Theorem~\ref{thm:main}}\label{sec:thmmainproof}
We note that the parameters of the services initialized by Algorithm~\ref{alg:initialize} are a subset of the users' unknown preferences. This allows us to define the notion of \textit{covering}. 
\begin{definition}
    Let $\Theta \subset \{\phi_1, \ldots, \phi_n\}$ be a set of services. A cluster $\mathcal{B} \in \mathcal{C}(\Theta_{\rm OPT})$ is said to be \textit{covered} by $\Theta$ if there exists $i \in \mathcal{B}$ such that $\phi_i \in \Theta$. If no such $i \in \mathcal{B}$ exists, then the cluster $\mathcal{B}$ is said to be \textit{uncovered}.
\end{definition}
The proof idea from here on is to show that there exists an approximation ratio $K_{\rm OPT}$ for the \textit{covered} clusters, and is shown in Lemma~\ref{lem:kmeans++new1} and \ref{lem:kmeans++new2}.
\begin{restatable}{lemma}{first}
\label{lem:kmeans++new1}
Let $\mathcal{B} \in \mathcal{C}(\Theta_{\rm OPT})$ be an arbitrary cluster in $\mathcal{C}(\Theta_{\rm OPT})$, and let $\theta \in \mathbb{R}^d$ be a service centered on the preference of a user $j$ chosen uniformly at random from $[n]$. The expected loss of users in $\mathcal{B}$, conditioned on $j \in \mathcal{B}$, satisfies 
\begin{align*}
    \mathbb{E}_{\theta}[\mathcal{L}(\{\theta = \phi_j\}, \mathcal{B}) \;|\;j \in \mathcal{B}] \leq \left(\max_{j \in \mathcal{B}} \frac{2}{|\mathcal{B}|}\sum_{i}\frac{1}{c_{ij}}\right) \mathcal{L}(\Theta_{\rm OPT}, \mathcal{B}).
\end{align*}
\end{restatable}
\begin{proof} 
Since we choose a user from $\mathcal{B}$, the conditional probability that we choose some fixed $\phi_j$ as the parameter for the service is precisely $P(\theta = \phi_j \;|\; \theta \in \mathcal{B}) = \frac{1}{|\mathcal{B}|}$. Let $\mathcal{J}(\mathcal{B})$ denote the best service covering all the points in $\mathcal{B}$, and then compute
\begin{align*}
    \mathbb{E}_{\theta}[\mathcal{L}(\{\theta = \phi_j\}, \mathcal{B}) \;|\;j \in \mathcal{B}]
    = \sum_{j \in \mathcal{B}} P(\theta = \phi_j) \mathcal{L}(\theta, \mathcal{B})
    = \sum_{j \in \mathcal{B}}\frac{1}{|\mathcal{B}|} \sum_{i \in \mathcal{B}}\mathcal{L}_i(\phi_j, \phi_i).
\end{align*}
Using Assumption~\ref{ass:triangle}.(ii) with $\theta = \mathcal{J}(\mathcal{B})$, we have that
\begin{align*}
\mathbb{E}_{\theta}[\mathcal{L}(\{\theta = \phi_j\}, \mathcal{B}) \;|\;j \in \mathcal{B}]
    &\leq \sum_{j \in \mathcal{B}} \sum_{i \in \mathcal{B}}\frac{1}{\vert \mathcal{B} \vert} \left(\frac{1}{c_{ij}}\left(\mathcal{L}_j(\mathcal{J}(\mathcal{B}), \phi_j) + \mathcal{L}_i(\mathcal{J}(\mathcal{B}), \phi_i)\right)\right), \\
    &= \sum_{j \in \mathcal{B}} \left(\frac{1}{|\mathcal{B}|}\sum_{i \in \mathcal{B}} \frac{1}{c_{ij}}\right) \mathcal{L}_j(\mathcal{J}(\mathcal{B}), \phi_j)  + \sum_{i \in \mathcal{B}} \left(\frac{1}{|\mathcal{B}|}\sum_{j \in \mathcal{B}} \frac{1}{c_{ij}}\right) \mathcal{L}_i(\mathcal{J}(\mathcal{B}), \phi_i).
\end{align*}
Noting that $c_{ij} = c_{ji}$, by swapping the indices of the second term in the above summand,  the second term is identical to the first term. Therefore, we deduce that
\begin{align*}
    \mathbb{E}_{\theta}[\mathcal{L}(\{\theta = \phi_j\}, \mathcal{B}) \;|\;j \in \mathcal{B}] \leq 2\sum_{j \in \mathcal{B}} \left(\left(\frac{1}{|\mathcal{B}|}\sum_{i \in \mathcal{B}} \frac{1}{c_{ij}}\right) \mathcal{L}_j(\mathcal{J}(\mathcal{B}), \phi_j)  \right).
\end{align*}
Now, using the fact that $\max_{j \in \mathcal{B}} \left(\frac{1}{|\mathcal{B}|}\sum_{i \in \mathcal{B}} \frac{1}{c_{ij}}\right)$
is a uniform upper bound for the multiplier in the above expression, we bring it outside the summation as a constant. 
Therefore, we deduce that
%Also, noting that , we further deduce that 
\begin{align}\label{eq:uniform_center}
    \mathbb{E}_{\theta}[\mathcal{L}(\{\theta = \phi_j\}, \mathcal{B}) \;|\;j \in \mathcal{B}] \leq \left(\max_{j \in \mathcal{B}} \frac{2}{|\mathcal{B}|}\sum_{i}\frac{1}{c_{ij}}\right)\mathcal{L}(\mathcal{J}(\mathcal{B}), \mathcal{B}),
\end{align}
where we have used the fact that $\mathcal{L}(\mathcal{J}(\mathcal{B}), \mathcal{B}) = \sum_{i \in \mathcal{B}}\mathcal{L}_i(\mathcal{J}(\mathcal{B}), \phi_i)$.
Since $\mathcal{B} \in \mathcal{C}(\Theta_{\rm OPT})$, the covering $\mathcal{J(B)}$ is the loss minimizing service among all services in $\Theta_{\rm OPT}$ for all points in $\mathcal{B}$. Thus, we have that
\begin{align*}
    \mathbb{E}_{\theta}[\mathcal{L}(\{\theta = \phi_j\}, \mathcal{B}) \;|\;j \in \mathcal{B}] \leq \left(\max_{j \in \mathcal{B}} \frac{2}{|\mathcal{B}|}\sum_{i}\frac{1}{c_{ij}}\right)\mathcal{L}(\Theta_{\rm OPT}, \mathcal{B}).
\end{align*}
This concludes the proof.
\end{proof}
\begin{restatable}{lemma}{subsequent}
\label{lem:kmeans++new2}
    Let $\mathcal{B} \in \mathcal{C}(\Theta_{\rm OPT})$  be an arbitrary cluster in $\mathcal{C}(\Theta_{\rm OPT})$, and let $\Theta_t \subset \mathbb{R}^d$ denote the parameters for a set of preexisting $t$ arbitrary services. 
    Consider a new clustering $\Theta_{t+1}=\Theta_t\cup\theta$ where $\theta = \phi_j$ is a random service  centered on user $j\in[n]$ selected with  probability $P(\theta = \phi_j) \propto \mathcal{L}_j(\Theta_t, \phi_j)$. 
   % 
   % Suppose this service $\theta$ is added to $\Theta_t$
    %
   % 
   % If we add a random service, parameterized by $\theta = \phi_j$, to $\Theta_t$ centered on a user $j$ chosen from $[n]$ with probability $P(\theta = \phi_j) \propto \mathcal{L}_j(\Theta_t, \phi_j)$ to get a new clustering $\Theta_{t+1} = \Theta_t \cup \theta$, then 
   Then, the expected loss of $\mathcal{B}$, conditioned on $j \in \mathcal{B}$, satisfies 
\begin{align*}
    \mathbb{E}_{\theta}[\mathcal{L}(\Theta_{t} \cup (\theta=\phi_j), \mathcal{B})\;|\;j \in \mathcal{B}]
    \leq \frac{1}{\min_{j \in \mathcal{B}}\frac{1}{|\mathcal{B}|}\sum_{i \in \mathcal{B}} c_{ji}}\left(\max_{j \in \mathcal{B}} \frac{2}{|\mathcal{B}|}\sum_{i}\frac{1}{c_{ij}}\right) \mathcal{L}(\Theta_{\rm OPT}, \mathcal{B}).
\end{align*}
\end{restatable}
\begin{proof}
    Given that we are choosing a user from $\mathcal{B}$, the conditional probability that we center the new service on some fixed $\phi_j$ is precisely $\mathcal{L}(\Theta_t, \phi_j)/(\sum_{i \in \mathcal{B}} \mathcal{L}(\Theta_t, \phi_i))$. After adding $\phi_j$ to the list of services, a user $i$ will have loss $\min\{\mathcal{L}_i(\Theta_t, \phi_i), \mathcal{L}_i(\phi_j,\phi_i)\}$.
Therefore we deduce that
\begin{equation}\label{eq:expected_lem4}
       \begin{aligned}
    \mathbb{E}_{\theta}[\mathcal{L}(\Theta_{t} \cup (\theta=\phi_j), \mathcal{B})\;|\;j \in \mathcal{B}] &= \sum_{j \in \mathcal{B}}P(\theta = \phi_j|\; \theta \in \mathcal{B})\sum_{i \in \mathcal{B}}  \mathcal{L}_i(\Theta_t \cup \theta, \phi_i),\\
   & = \sum_{j \in \mathcal{B}} \frac{\mathcal{L}_j(\Theta_t, \phi_j)}{\sum_{l \in \mathcal{B}}\mathcal{L}_l(\Theta_t, \phi_l)}\sum_{i \in \mathcal{B}}  \min\{\mathcal{L}_i(\Theta_t, \phi_i), \mathcal{L}_i(\phi_j,\phi_i)\}.
    \end{aligned}
\end{equation}
By Assumption~\ref{ass:triangle}.$(i)$, for any $\theta \in \mathbb{R}^d$ we have that
\begin{align*}
    c_{ji} \mathcal{L}_j(\theta, \phi_j) \leq \mathcal{L}_i(\theta, \phi_i) + \mathcal{L}_i(\phi_j, \phi_i).
\end{align*}
We utilize the fact that given two functions $f,g: \Theta \rightarrow \mathbb{R}$, if $f(\theta) \leq g(\theta) \;\forall \theta \in \Theta$, then $\min_{\theta \in \Theta} f(\theta) \leq \min_{\theta \in \Theta} g(\theta)$. Hence, the following implication holds: 
\begin{align*}
    c_{ji} \min_{\theta \in \Theta}\mathcal{L}_j(\theta, \phi_j) \leq \min_{\theta \in \Theta} \mathcal{L}_i(\theta, \phi_i) + \mathcal{L}_i(\phi_j, \phi_i)\  \implies \ c_{ji} \mathcal{L}_j(\Theta_t, \phi_j) \leq \mathcal{L}_i(\Theta_t, \phi_i) + \mathcal{L}_i(\phi_j, \phi_i).
\end{align*}
Summing over all $i \in \mathcal{B}$, we get that
\begin{align*}
    \mathcal{L}_j(\Theta_t, \phi_j)\leq \frac{1}{\sum_{i \in \mathcal{B}} c_{ji}} \left(\sum_{i \in \mathcal{B}} \mathcal{L}_i(\Theta_t, \phi_i) + \mathcal{L}_i(\phi_j, \phi_i)\right)
\end{align*}
 Applying this to $\mathcal{L}_j(\Theta_t, \phi_j)$ on the right hand side of  \eqref{eq:expected_lem4}, we have that
\begin{align*}
      \mathbb{E}_{\theta}[\mathcal{L}(\Theta_{t} \cup (\theta=\phi_j), \mathcal{B})\;|\;j \in \mathcal{B}] &\leq \sum_{j \in \mathcal{B}}\frac{1}{\sum_{i \in \mathcal{B}} c_{ji}}\left(1 +  \frac{\sum_{i \in \mathcal{B}}\mathcal{L}_i(\phi_j, \phi_i)}{\sum_{l \in \mathcal{B}}\mathcal{L}_l(\Theta_t, \phi_l)}\right)\sum_{i \in \mathcal{B}} \min\{\mathcal{L}_i(\Theta_t, \phi_i), \mathcal{L}_i(\phi_j,\phi_i)\}.
\end{align*}
Since $\min\{a,b\}\leq a$, we further deduce that
\begin{align*}
      \mathbb{E}_{\theta}[\mathcal{L}(\Theta_{t} \cup (\theta=\phi_j), \mathcal{B})\;|\;j \in \mathcal{B}] &\leq \sum_{j \in \mathcal{B}}\frac{1}{\sum_{i \in \mathcal{B}} c_{ji}} \sum_{i \in \mathcal{B}}\mathcal{L}_i(\phi_j,\phi_i)\\
      &\qquad+ \sum_{j \in \mathcal{B}}\frac{1}{\sum_{i \in \mathcal{B}} c_{ji}}\frac{\sum_{i \in \mathcal{B}}\mathcal{L}_i(\phi_j,\phi_i)}{\sum_{l \in \mathcal{B}}\mathcal{L}_l(\Theta_t, \phi_l)}\sum_{i \in \mathcal{B}} \mathcal{L}(\Theta_t, \phi_i) ,\\
      &=\sum_{j \in \mathcal{B}}\frac{2}{\sum_{i \in \mathcal{B}} c_{ij}} \sum_{i \in \mathcal{B}}\mathcal{L}_i(\phi_j,\phi_i).
\end{align*}
Now, we use the upper bound $\frac{1}{\min_{j \in \mathcal{B}} \sum_{i \in \mathcal{B}}c_{ji}}$ for the multiplier 
to get
\begin{align*}
    \mathbb{E}_{\theta}[\mathcal{L}(\Theta_{t} \cup (\theta=\phi_j), \mathcal{B})\;|\;j \in \mathcal{B}] \leq \frac{2}{\frac{1}{|\mathcal{B}|}\min_{j \in \mathcal{B}} \sum_{i \in \mathcal{B}}c_{ji}} \sum_{j \in \mathcal{B}} \sum_{i \in \mathcal{B}}\frac{1}{|\mathcal{B}|}\mathcal{L}_i(\phi_j,\phi_i).
\end{align*}
Note that $\sum_{j \in \mathcal{B}} \sum_{i \in \mathcal{B}}\frac{1}{|\mathcal{B}|}\mathcal{L}_i(\phi_j,\phi_i)$ is essentially the loss on choosing users from $\mathcal{B}$ uniformly randomly and centering the service on the chosen user. Plugging in the expression in  \eqref{eq:uniform_center} from the proof of Lemma \ref{lem:kmeans++new1}, we have that 
\begin{align}\label{eq:iterative_center}
   \mathbb{E}_{\theta}[\mathcal{L}(\Theta_{t} \cup (\theta=\phi_j), \mathcal{B})\;|\;j \in \mathcal{B}] \leq \frac{4}{\min\limits_{j \in \mathcal{B}}\frac{1}{|\mathcal{B}|}\sum \limits_{i \in \mathcal{B}} c_{ji}}\left(\max_{j \in \mathcal{B}} \frac{1}{|\mathcal{B}|}\sum_{i \in \mathcal{B}}\frac{1}{c_{ij}}\right) \mathcal{L}(\mathcal{J}(\mathcal{B}), \mathcal{B}).
\end{align}
Since $\mathcal{B} \in \mathcal{C}(\Theta_{\rm OPT})$, the covering $\mathcal{J(B)}$ is the loss minimizing service among all available services in $\Theta_{\rm OPT}$ for all points in $\mathcal{B}$. Therefore, we deduce that
\begin{align*}
    \mathbb{E}_{\theta}[\mathcal{L}(\Theta_{t} \cup (\theta=\phi_j), \mathcal{B})\;|\;j \in \mathcal{B}] \leq \frac{4}{\min\limits_{j \in \mathcal{B}}\frac{1}{|\mathcal{B}|}\sum \limits_{i \in \mathcal{B}} c_{ji}}\left(\max_{j \in \mathcal{B}} \frac{1}{|\mathcal{B}|}\sum_{i \in \mathcal{B}}\frac{1}{c_{ij}}\right) \mathcal{L}(\Theta_{\rm OPT}, \mathcal{B}),
\end{align*}
which concludes the proof.
\end{proof}
The following Lemma is an induction relating the losses on \textit{covered} and \textit{uncovered} clusters.
\begin{restatable}{lemma}{induction}
\label{lem:induction}
    Let $\Theta_t \subset \mathbb{R}^d$ denote the parameters of a set of $t$ arbitrary services. Consider  $u > 0$ \emph{uncovered} clusters from $\mathcal{C}(\Theta_{\rm OPT})$, denoted by $\mathcal{U}_t$, and let $\mathcal{H}_t$ denote the covered clusters. Suppose we add $v \leq u$ random services to $\Theta_t$, chosen with probability proportional to their current loss---i.e., $\mathcal{L}(\Theta_t, \phi_j)$---and let $\Theta_{t + v}$ denote the the resulting set of services. The following estimate holds:
    \begin{align*}
        \mathbb{E}_{\Theta_{t + v}}[\mathcal{L}(\Theta_{t + v}, [n])] \leq \left(\mathbb{E}_{\Theta_t}[\mathcal{L}\left(\Theta_t, \mathcal{H}_t\right)]+ K_{\rm OPT}\cdot \mathcal{L}\left(\Theta_{\rm OPT}, \mathcal{U}_t\right)\right)\cdot\left(1 + S_v\right) + \frac{u - v}{u}\cdot  \mathbb{E}_{\Theta_t}[\mathcal{L}\left(\Theta_t, \mathcal{U}_t\right)],
    \end{align*}
    where $K_{\rm OPT}$ is as defined in \eqref{eqn:K} and $S_v = \left(1 + \frac{1}{2} + \ldots + \frac{1}{v}\right)$ is the harmonic series.
\end{restatable}
\begin{proof}
    Replacing the factor 8 in Lemma 3.3 from \citep{arthur2007k} by $K_{\rm OPT}$ gives us the desired result.
    
\end{proof}
With the preceding technical lemmas, we are now ready to prove the main theorem which we restate for convenience. 
\main*
\begin{proof}
    Consider $t=1$, and $\mathcal{B} \in \mathcal{C}(\Theta_{\rm OPT})$, the cluster in which the first chosen user belongs. Applying Lemma \ref{lem:induction} with $v = u = k-1$ and the fact that $\mathcal{B}$ is the only covered cluster, we have that
    \begin{align*}
        \mathbb{E}_{\Theta_k}[\mathcal{L}(\Theta_{k}, [n])] \leq \left(\mathbb{E}_{\Theta_1}[\mathcal{L}\left(\Theta_1, \mathcal{B}\right)] + K_{\rm OPT}\cdot \mathcal{L}\left(\Theta_{\rm OPT}, [n] - \mathcal{B}\right)\right)\cdot\left(1 + S_{k-1}\right).
    \end{align*}
 Observe that
    \begin{align*}
        \mathbb{E}_{\Theta_1}[\mathcal{L}\left(\Theta_1, \mathcal{B}\right)] &\leq K_{\rm OPT} \cdot\mathcal{L}\left(\Theta_{\rm OPT}, \mathcal{B}\right)],
    \end{align*} 
    by Lemma \ref{lem:kmeans++new1}. Moreover, we have that
    \[ \mathcal{L}\left(\Theta_{\rm OPT}, [n] - \mathcal{B}\right) = \mathcal{L}\left(\Theta_{\rm OPT}, [n]\right) - \mathcal{L}\left(\Theta_{\rm OPT}, \mathcal{B}\right).\]
    Using these two expression and noting $S_{k-1} \leq 1 + \log k$, we get the stated result. 
\end{proof}

\section{Fair Initialization}\label{sec:fair}
We assume the scenario when the user demographic identities are known \emph{a priori}. Let $\gamma : [n] \rightarrow \mathcal{A}$ be a mapping, which maps a user $i$ to its demographic group $\gamma(i)$, and let $|\gamma(i)|$ denote the size of the demographic group user $i$ belongs to. 

We define the weighted total loss (where each users' loss is divided by its group size) as: 
\begin{align*}
    \mathcal{G}(\Theta, [n]) &= \sum_{i \in [n]} \frac{\mathcal{L}_i(\Theta, \phi_i)}{|\gamma(i)|}= \sum_{i \in [m]} \frac{\mathcal{L}(\Theta, \mathcal{A}_i)}{|\mathcal{A}_i|}.
\end{align*}
Note that this can also be interpreted as the sum of average losses across different demographic groups. 

We introduce Algorithm~\ref{alg:fair_initialize} with a couple of changes from Algorithm~\ref{alg:initialize}.
\begin{itemize}[itemsep=-2pt]
    \item Instead of uniformly picking a user at random, we are picking a user $i$ with probability proportional to $\frac{1}{|\gamma(i)|}$. This is equivalent to saying, the probability of the first user being in some group $\mathcal{A}_j \in \mathcal{A}$ is \[\frac{\sum_{i \in \mathcal{A}_j} \frac{1}{|\gamma(i)|}}{\sum_{j \in [m]}\sum_{i \in \mathcal{A}_j} \frac{1}{|\gamma(i)|}} = \frac{1}{m}.\] That is, the first user is equally likely to belong to one of the $m$ groups.
    \item A user $i$ at any time step $t$ is selected with probability proportional to $\mathcal{L}_i(\Theta_{t-1}, \phi_i) / |\gamma(i)|$. This implies that a user from some group $\mathcal{A}_j \in \mathcal{A}$ is likely to be picked with probability proportional to the average demographic loss---i.e., \[\sum_{i \in \mathcal{A}_j} \mathcal{L}_i(\Theta_{t-1}, \phi_i) / |\gamma(i)| = \frac{\mathcal{L}(\Theta_{t-1}, \mathcal{A}_j)}{|\mathcal{A}_j|}.\]
\end{itemize}
We first state the approximation ratio of Algorithm~\ref{alg:fair_initialize} on the objective $\mathcal{G}(\Theta, [n])$.
\begin{lemma}
\label{lem:main_rescaled}
    Consider $n$ users with unknown preferences $\{\phi_1, \ldots, \phi_n\} \subset \mathbb{R}^d$, and unknown associated loss functions  $\mathcal{L}_i(\cdot,\cdot)$ satisfying Assumptions \ref{ass:unique} and \ref{ass:triangle}. Let $\Theta_{\rm scaled} \subset \mathbb{R}^d$ be the set of $k$ services minimizing the total loss and $\mathcal{C}(\Theta_{\rm scaled})$ the resulting partitioning of users. If Algorithm \ref{alg:fair_initialize} is used to obtain $k$ services $\Theta_k$, then the following bound holds:
    \begin{align*}
        \mathbb{E}_{\Theta_k}[\mathcal{G}(\Theta_k, [n])] &\leq  K_{\rm fair}(2 + \log k) \cdot \mathcal{G}(\Theta_{\rm scaled}, [n]),
    \end{align*}
    where the expectation is taken over the randomization of the algorithm and $K_{\rm fair}$ is equal to 
    % \begin{equation}\label{eqn:K_scaled}
    %    \max_{\mathcal{B} \in \mathcal{C}(\Theta_{\rm scaled})}
    %    \frac{4}{\min\limits_{j \in \mathcal{B}}\frac{1}{|\mathcal{B}|}\sum \limits_{i \in \mathcal{B}} \frac{c_{ij}}{|\gamma(i)|}}\left(\max_{j \in \mathcal{B}} \frac{1}{|\mathcal{B}|}\sum_{i \in \mathcal{B}}\frac{1}{c_{ij}|\gamma(i)|}\right).
    % \end{equation}
    \begin{equation}\label{eqn:K_scaled}
       \max_{\mathcal{B} \in \mathcal{C}(\Theta_{\rm scaled})}
       \frac{4}{\min\limits_{j \in \mathcal{B}}\sum \limits_{i \in \mathcal{B}} \frac{c_{ij}}{|\gamma(i)|}}\left(\max_{j \in \mathcal{B}} \sum_{i \in \mathcal{B}}\frac{1}{c_{ij}|\gamma(i)|}\right).
    \end{equation}
\end{lemma}
    The proof is similar as Theorem~\ref{thm:main}, with the introduction of $\frac{1}{|\gamma(i)|}$'s when deriving Lemma~\ref{lem:kmeans++new1} and \ref{lem:kmeans++new2}.

\begin{algorithm}[t!]
\caption{Fair AcQUIre- Fair Adaptively Querying Users for Initialization}
\begin{algorithmic}[1]
\STATE \textbf{Input:} Set of users $[n]$, number of services $k$, Demographic groups $\mathcal{A} = \{\mathcal{A}_1, \ldots, \mathcal{A}_m\}$,  Map users to demographic groups $\gamma : [n] \rightarrow \mathcal{A}$
\STATE Choose a user $i$ uniformly randomly from $[n]$ with probability $\propto \frac{1}{|\gamma(i)|}$.
\STATE Query user $i$'s preference $\phi_i$, set the first service $\Theta_1 = \phi_i$.
\FOR{$t \in \{2, \ldots, k\}$}
        \STATE \textbf{User behavior:} Collect user losses on existing services $\Theta_{t-1}$ : \{$\mathcal{L}_i(\Theta_{t-1}, \phi_i)\}_{i \in [n]}$.
        \STATE \textbf{User Selection:} Sample $l$ from $[n]$ with probability $P(l = i) \propto \mathcal{L}_i(\Theta_{t-1}, \phi_i) / |\gamma(i)|$.
        \STATE \textbf{New service:} Query user $l$'s preference $\phi_l$.
        \STATE $\Theta_t = \Theta_{t-1} \cup \phi_l$.
\ENDFOR
\STATE \textbf{Return} $\Theta_k$ 
\end{algorithmic}
\label{alg:fair_initialize}
\end{algorithm}
% \begin{theorem}
% \label{thm:fair_known_groups}
%     Consider $n$ users with unknown preferences $\{\phi_1, \ldots, \phi_n\} \subset \mathbb{R}^d$, and unknown associated loss functions  $\mathcal{L}_i(\cdot,\cdot)$ satisfying Assumptions \ref{ass:unique} and \ref{ass:triangle}. Suppose these users belong to $m$ demographic groups $\mathcal{A} = \{\mathcal{A}_1, \ldots, \mathcal{A}_m\} \subset [n]$. Let $\Theta_{\rm fair} \subset \mathbb{R}^d$ be the set of $k$ services minimizing the fairness objective $\Phi$ given in \eqref{eq:fair}. If Algorithm \ref{alg:fair_initialize} is used to obtain $k$ services $\Theta_k$, then the following bound holds:
%     \begin{align*}
%          &\mathbb{E}_{\Theta_k}[\Phi(\Theta_k, \mathcal{A})] \leq m K_{\rm fair} (2 + \log k) \cdot \Phi(\Theta_{\rm fair}, \mathcal{A}),
%     \end{align*}
%     where the expectation is taken over the randomization of the algorithm and $K_{\rm fair}$ is defined in  \eqref{eqn:K_scaled}.
% \end{theorem}
\fairknowngroups*
\begin{proof}
For any $\Theta \subset \mathbb{R}^d$, we have that
\begin{align*}
    \Phi(\Theta, \mathcal{A}) &= \max_{i \in [m]} \frac{\mathcal{L}(\Theta, \mathcal{A}_i)}{|\mathcal{A}_i|}\leq \sum_{i \in [m]} \frac{\mathcal{L}(\Theta, \mathcal{A}_i)}{|\mathcal{A}_i|}=\mathcal{G}(\Theta, [n]).
\end{align*}
For $\Theta_k$---namely the output of Algorithm~\ref{alg:initialize}---the above expression implies that
\begin{align*}
    \mathbb{E}_{\Theta_k}[\Phi(\Theta_k, \mathcal{A})] \leq \mathbb{E}_{\Theta_k}[\mathcal{G}(\Theta_k, [n])].
\end{align*}
We also have that
\begin{align*}
\mathcal{G}(\Theta_{\rm fair}, [n]) &= \sum_{i \in [m]} \frac{\mathcal{L}(\Theta_{\rm fair}, \mathcal{A}_i)}{|\mathcal{A}_i|}\leq \sum_{i \in [m]}\max_{i \in [m]} \frac{\mathcal{L}(\Theta_{\rm fair}, \mathcal{A}_i)}{|\mathcal{A}_i|}= m\cdot\Phi(\Theta_{\rm fair}, \mathcal{A}).
\end{align*}    
Let $\Theta_{\rm scaled}$ be the minimizer of $\mathcal{G}(\cdot, [n])$ and $\Theta_{\rm fair}$ be the minimizer of $\Phi(\cdot, \mathcal{A})$. Then, we have that
\begin{align*}
    \mathcal{G}(\Theta_{\rm scaled}, [n]) \leq \mathcal{G}(\Theta_{\rm fair}, [n]) \leq m\cdot\Phi(\Theta_{\rm fair}, \mathcal{A}) \leq  m\cdot \mathbb{E}_{\Theta_k}[\Phi(\Theta_k, \mathcal{A})] \leq m \cdot \mathbb{E}_{\Theta_k}[\mathcal{G}(\Theta_k, [n])].
\end{align*}
Now using Lemma~\ref{lem:main_rescaled} for the guarantee on $\Theta_k$ for $\mathcal{G}(\cdot, [n])$, we conclude the proof. 
\end{proof}
\begin{remark}
    If all $c_{ij}$'s are identically equal to some $c > 0$, $K_{\rm fair} = \frac{4}{c^2}$ and the approximation ratio of Algorithm~\ref{alg:fair_initialize} for the fair objective is $4m(2 + \log k)/ c^2$. Meanwhile, Algorithm~\ref{alg:initialize} would have an approximation ratio \[4m\cdot\frac{\max_{i \in [m]} |\mathcal{A}_i|}{\min_{i \in [m]} |\mathcal{A}_i|}\cdot\frac{(2 + \log k)}{c^2}.\]
\end{remark}
\section{Proof of Theorem~\ref{thm:regression}} \label{sec:finite_sample}
Let the empirical loss---which is a finite sample average regression loss---be denoted 
\[\widehat{\mathcal{L}}_i(\theta, \phi_i) = \frac{1}{n_i}\sum_{j \in [n_i]}(\theta^\top x_i^j - y_i^j)^2,\] where
 $\{(x_i^j, y_i^j)\}_{j \in [n_i]}$ are \textbf{private unknown} features and scores of the users, respectively.
\begin{lemma}\label{lem:geneigenvalue}
Under Assumptions~\ref{ass:independent} and \ref{ass:fullrank}, each subpopulation empirical loss $\widehat{\mathcal{L}}_i$
 satisfies Assumption~\ref{ass:unique} and Assumption~\ref{ass:triangle} with
\begin{align*}
c_{ij} = \frac{1}{2}\lambda_{\mathrm{min}}\left(\frac{\sum_{l \in [n_i]}x_i^l(x_i^l)^\top}{n_i}, \frac{\sum_{l \in [n_j]}x_j^l(x_j^l)^\top}{n_j}\right).
\end{align*}
\end{lemma}
\begin{proof}
Fix a subpopulation index $i$. We start by showing $\widehat{\mathcal{L}}_i$ satisfies Assumption~\ref{ass:unique}. We use the following result: if $p \leq d$ random vectors in $\mathbb{R}^d$ are independently drawn from a distribution that is absolutely continuous with respect to the Lebesgue measure, then they are almost surely linearly independent. 
By Assumption~\ref{ass:fullrank}, applying this result to our scenario, if we draw $n_i \geq d$ features independently from a sub-Gaussian distribution to form a feature matrix $\mathbf{X}_i \in \mathbb{R}^{n_i \times d}$, then it is almost surely full column rank. 

Therefore, we compute $\phi_i = \mathbf{X}_i^\dagger y_i$ since $\mathbf{X}_i^\dagger \mathbf{X}_i = \mathbf{I}_p$ when $\mathbf{X}_i$ is full column rank. The subpopulation $i$'s empirical loss is compactly written as $\widehat{\mathcal{L}}_i(\theta, \phi_i) = \|\theta - \phi_i\|_{\mathbf{A}_i}^2$, where $\mathbf{A}_i = \frac{1}{n_i} \sum_{l \in [n_i]} x_i^l(x_i^l)^\top$. Thus $\widehat{\mathcal{L}}_i(\theta, \phi_i)$ satisfies Assumption~\ref{ass:unique}.

We now show that $\widehat{\mathcal{L}}_i$  satisfies Assumption~\ref{ass:triangle} with $c_{ij} = \frac{1}{2}\lambda_{\rm \min}(\mathbf{A}_i, \mathbf{A}_j)$.
To this end, we find the largest $c\in \mathbb{R}_{+}$ such that $c  (u^\top\mathbf{A}_i u)\leq  
u^\top\mathbf{A}_j u$ and $c  (u^\top\mathbf{A}_j u)\leq  
u^\top\mathbf{A}_i u$ for all $u\in \mathbb{R}^d$.

Rearranging the inequality, this problem is the same as finding the largest $c\in \mathbb{R}_{+}$ such that
\[u^\top ( 
\mathbf{A}_i- c\mathbf{A}_j)u\geq 0\quad \text{and} \quad \mathbf{A}_j- c\mathbf{A}_i)u\geq 0 \quad \forall \ u\in \mathbb{R}^d,\]
or equivalently,
\[ 
\mathbf{A}_i-c\mathbf{A}_j\succeq 0 \quad \text{and} \quad \mathbf{A}_j-c\mathbf{A}_i\succeq 0.\]
Therefore, finding such a constant $c$ is easily  reformulated as the following optimization problem:
\begin{align*}
    \max\{c ~|\ 
   \mathbf{A}_i - c\mathbf{A}_j \succeq 0, \mathbf{A}_j - c\mathbf{A}_i \succeq 0\} =\lambda_{\rm \min}(\mathbf{A}_i, \mathbf{A}_j).
\end{align*}

There are two key tools we use to finish the argument.
\begin{itemize}[itemsep=-1pt]
    \item For any $u \in \mathbb{R}^d$, $\frac{1}{2}\lambda_{\rm \min}(\mathbf{A}_i, \mathbf{A}_j) \|u\|_{\mathbf{A}_i}^2 \leq \|u\|_{\mathbf{A}_j}^2$.
    \item For any norm triangle inequality gives us $\|\theta - \phi_i\| \leq \|\theta - \phi_j\| + \|\phi_i - \phi_j\|$. Then we use power mean inequality, i.e. $a \leq b + c \implies a^2 \leq 2(b^2 + c^2)$.
\end{itemize}
The following shows Assumption~\ref{ass:triangle}.$(i)$ holds with $c_{ij} = \frac{1}{2}\lambda_{\rm \min}(\mathbf{A}_i, \mathbf{A}_j)$:
\begin{align*}
    \frac{1}{2}\lambda_{\rm \min}(\mathbf{A}_i, \mathbf{A}_j)\widehat{\mathcal{L}}_i(\theta, \phi_i)
    &= \frac{1}{2}\lambda_{\rm \min}(\mathbf{A}_i, \mathbf{A}_j) \|\theta - \phi_i\|_{\mathbf{A}_i}^2\\
    &\leq \frac{1}{2}\|\theta - \phi_i\|_{\mathbf{A}_j}^2\\
    &\leq \|\theta - \phi_j\|^2_{\mathbf{A}_j} + \|\phi_i - \phi_j\|^2_{\mathbf{A}_j}\\
    &= \widehat{\mathcal{L}}_j(\theta, \phi_j) + \widehat{\mathcal{L}}_j(\phi_i, \phi_j).
\end{align*}
Analogously, the following shows Assumption~\ref{ass:triangle}.$(ii)$ holds with $c_{ij} = \frac{1}{2}\lambda_{\rm \min}(\mathbf{A}_i, \mathbf{A}_j)$:
\begin{align*}
    \frac{1}{2}\lambda_{\rm \min}(\mathbf{A}_i, \mathbf{A}_j)\widehat{\mathcal{L}}_i(\phi_j, \phi_i)
    &= \frac{1}{2}\lambda_{\rm \min}(\mathbf{A}_i, \mathbf{A}_j) \|\phi_j - \phi_i\|_{\mathbf{A}_i}^2\\
    &\leq \lambda_{\rm \min}(\mathbf{A}_i, \mathbf{A}_j) \left(\|\theta - \phi_j\|_{\mathbf{A}_i}^2 +  \|\theta - \phi_i\|_{\mathbf{A}_i}^2\right)\\
    &\leq \|\theta - \phi_j\|_{\mathbf{A}_j}^2 + \|\theta - \phi_i\|^2_{\mathbf{A}_i}\\
    &= \mathcal{L}_j(\theta, \phi_j) + \mathcal{L}_i(\theta, \phi_i).
\end{align*}
This concludes the proof.
\end{proof}
With the preceding technical lemma in place, we now are ready to prove Theorem~\ref{thm:regression}.
\regression*

\begin{proof}
Lemma~\ref{lem:geneigenvalue} gives us that the subpopulation $i$'s empirical loss for a regressor $\theta \in \mathbb{R}^d$ can be written as $\|\theta - \phi_i\|^2_{\mathbf{A}_i}$. This is a random quantity since $\mathbf{A}_i$ is a sample average covariance of randomly chosen features. We analyse this term next.

For a random feature $x \in \mathbb{R}^d$ in subpopulation $i$, given a fixed center $\theta \in \mathbb{R}^d$, the term $\frac{(\theta - \phi_i)^\top x}{\|\theta - \phi_i\|_{\Sigma_i}}$ is sub-Gaussian with variance proxy $\sigma_i^2$ by Assumption~\ref{ass:independent}. The square of a sub-Gaussian random variable,  $\frac{(\theta - \phi_i)^\top xx^\top(\theta - \phi_i)}{\|\theta - \phi_i\|^2_{\Sigma_i}}$ is sub-exponential\footnote{A random variable $x$ is subexponential with parameters $(v^2, \alpha)$ if $\mathbb{E}[\exp(\lambda x)] \leq \exp(\frac{\nu^2 \lambda^2}{2}) \; \forall |\lambda| \leq \frac{1}{\alpha}$.} with parameters $\nu_i^2 = 32\sigma_i^4, \alpha_i=4\sigma_i^2$ (see \citep{honorio2014tight}[Appendix B]). 

Since subpopulation $i$'s empirical loss is an average of $n_i$ samples, the empirical loss $\|\theta - \phi_i\|^2_{\mathbf{A}_i}$ is a subexponential variable with parameters $\frac{\nu_i^2}{n_i}\|\theta - \phi_i\|^4_{\Sigma_i}, \frac{\alpha_i}{\sqrt{n_i}}\|\theta - \phi_i\|^2_{\Sigma_i}$. This is because the variance of sample average scales as $\frac{1}{\sqrt{n_i}}$, and the constant $\|\theta - \phi_i\|^2_{\Sigma_i}$ scales the subexponential parameters appropriately. 

Now, given a set of $k$ arbitrary services $\Theta = \{\theta_1, \ldots, \theta_k\} \subset \mathbb{R}^d$, let $\theta_{\gamma(i)} \in \Theta$ be an arbitrary service chosen by subpopulation $i$. The sum of empirical losses on based on such choices is also subexponential with parameters $\nu^2 = \sum_{i \in [N]}\frac{\nu_i^2}{n_i}\|\theta - \phi_i\|^4_{\Sigma_i}$ and $\alpha = \max_{i \in [N]} \frac{\alpha_i}{\sqrt{n_i}}\|\theta - \phi_i\|^2_{\Sigma_i}$. Using Chernoff bound \citep{vershynin2018high}, we have that
\begin{align}\label{eq:Chernoff}
    P\left\{|\sum_{i \in [N]}\|\theta_{\gamma(i)} - \phi_i\|^2_{\mathbf{A}_i} - \sum_{i \in [N]}\|\theta_{\gamma(i)} - \phi_i\|^2_{\Sigma_i}|\right\} \geq t] \leq 2\exp\left(-\frac{t^2}{2\nu^2}\right) \quad\text{where}\  0 \leq t \leq \frac{\nu^2}{\alpha}.
\end{align}
Note that the probability bound is is increasing in $\nu$, we thus take an upper bound on $\nu^2$ to upper bound this probability:
\begin{align*}
    \nu^2 &= \sum_{i \in [N]}\frac{\nu_i^2}{n_i}\|\theta - \phi_i\|^4_{\Sigma_i}= 32 \sum_{i \in [N]}\frac{\sigma_i^4}{n_i}  \|\theta_{\gamma(i)} - \phi_i\|^4_{\Sigma_i} \leq 32 \left(\sum_{i \in [N]} \frac{\sigma_i^8}{n_i^2}\right)^{\frac{1}{2}} \left(\sum_{i \in [N]} \|\theta_{\gamma(i)} - \phi_i\|^8_{\Sigma_i}\right)^{\frac{1}{2}},
\end{align*}
where the last inequality follows from Cauchy-Schwarz. Further, noticing that $\sum_{i \in n} a_i^4 \leq (\sum_{i \in n} a_i^2)^2$, and applying it for the last term twice, we deduce that
\begin{align*}
    \nu^2 \leq 32 \left(\sum_{i \in [N]} \frac{\sigma_i^8}{n_i^2}\right)^{\frac{1}{2}} \left(\sum_{i \in [n]} \|\theta_{\gamma(i)} - \phi_i\|^2_{\Sigma_i}\right)^2.
\end{align*}
Choosing $t = \epsilon \sum_{i \in [N]}\|\theta_{\gamma(i)} - \phi_i\|^2_{\Sigma_i}$ in \eqref{eq:Chernoff}, where $0 \leq \epsilon \leq \mathcal{O}(\min_{i \in [N]}\frac{\sigma_i^2}{\sqrt{n_i}})$, we have that
\begin{align*}
    &P\left[|\sum_{i \in [N]}\|\theta_{\gamma(i)} - \phi_i\|^2_{\mathbf{A}_i} - \sum_{i \in [N]}\|\theta_{\gamma(i)} - \phi_i\|^2_{\Sigma_i}| \geq \epsilon \sum_{i \in [N]}\|\theta_{\gamma(i)} - \phi_i\|^2_{\Sigma_i}\right]\\
    &\qquad\geq 2\exp{\left(-\frac{\epsilon^2}{64} \left({\sum_{i\in [N]} \frac{\sigma_i^8}{n_i^2}}\right)^{-1/2}\right)}.
\end{align*}
With high probability, at least $1 - 2\exp{\left(-\frac{\epsilon^2}{64} \left({\sum_{i\in [N]} \frac{\sigma_i^8}{n_i^2}}\right)^{-1/2}\right)}$, for all $\Theta \subset \mathbb{R}^d$, we have that
\begin{align*}
    (1 - \epsilon) \sum_{i \in [N]}\|\theta_{\gamma(i)} - \phi_i\|^2_{\Sigma_i} \leq \sum_{i \in [N]}\|\theta_{\gamma(i)} - \phi_i\|^2_{\mathbf{A}_i} \leq (1 + \epsilon)\sum_{i \in [N]}\|\theta_{\gamma(i)} - \phi_i\|^2_{\Sigma_i}.
\end{align*}
The rest of the proof is deterministic, thus it is assumed everything follows with probability at least $1 - 2\exp{\left(-\frac{\epsilon^2}{64} \left({\sum_{i\in [N]} \frac{\sigma_i^8}{n_i^2}}\right)^{-1/2}\right)}$. %
Define $\Theta^N = \Theta \times \cdots \times \Theta$---i.e., Cartesian product of the set $\Theta$ $N$-times. Given two functions $f,g: \Theta^N \rightarrow \mathbb{R}$, if $f(\theta) \leq g(\theta)$ for all $\theta \in \Theta^N$, then $\min_{\theta \in \Theta^N} f(\theta) \leq \min_{\theta \in \Theta^N} g(\theta)$. Therefore, we deduce that
\begin{align}\label{eq:finite_quadratic}
(1 - \epsilon) \sum_{i \in [N]}\min_{j \in [k]}\|\theta_j - \phi_i\|^2_{\Sigma_i} &\leq \sum_{i \in [N]}\min_{j \in [k]}\|\theta_j - \phi_i\|^2_{\mathbf{A}_i} \leq (1 + \epsilon)\sum_{i \in [N]}\min_{j \in [k]}\|\theta_j - \phi_i\|^2_{\Sigma_i}.
\end{align}
Hence, given the same set of services $\Theta$, the total empirical loss can be bounded with respect to the total expected loss as follows:
\begin{align*}
(1 - \epsilon) \mathcal{L}(\Theta, [N]) \leq  \widehat{\mathcal{L}}(\Theta, [N]) \leq (1 + \epsilon) \mathcal{L}(\Theta, [N]). 
\end{align*}
Recall $\Theta_{\rm OPT}$ denotes the optimal solution for the \textit{total expected loss} and let $\widehat{\Theta}_{\rm OPT}$ denote the optimal solution for the \textit{total empirical loss}. Let $\Theta_k$ be the output 
of Algorithm~\ref{alg:initialize} on finite samples per subpopulations.
Using the first inequality of \eqref{eq:finite_quadratic}, we have that
\begin{align}\label{eq:finite_left}
    (1 - \epsilon) \mathbb{E}_{\Theta_k}[\mathcal{L}(\Theta_k, [N])]\leq \mathbb{E}_{\Theta_k}[\widehat{\mathcal{L}}(\Theta_k, [N])].
\end{align}
Also, noting that $\widehat{\Theta}_{\rm OPT}$ is the minimizer for $\widehat{\mathcal{L}}(\Theta, [N])$ and using the second inequality of  \eqref{eq:finite_quadratic}, we get that
\begin{align}\label{eq:finite_right}
    \widehat{\mathcal{L}}(\widehat{\Theta}_{\rm OPT}, [N]) \leq \widehat{\mathcal{L}}(\Theta_{\rm OPT}, \Phi) \leq (1 + \epsilon) \mathcal{L}(\Theta_{\rm OPT}, \Phi). 
\end{align}
Now combining \eqref{eq:finite_left} and \eqref{eq:finite_right} with Theorem \ref{thm:main}, we get the desired result---i.e.,
\begin{align*}
    \mathbb{E}_{\Theta_k}[\mathcal{L}(\Theta_k, [N])] \leq \frac{1 + \epsilon}{1 - \epsilon} K_{\rm OPT}(2 + \log k)\cdot \mathcal{L}(\Theta_{\rm OPT}, [N]).
\end{align*}
Setting $\delta=1 - 2\exp{\left(-\frac{\epsilon^2}{64} \left({\sum_{i\in [N]} \frac{\sigma_i^8}{n_i^2}}\right)^{-1/2}\right)}$, we get that $n_i = \Omega(\frac{\sigma_i^4 \sqrt{N} \log{(2/\delta)}}{\epsilon^2})$. This concludes the proof.

\end{proof}
 
\begin{figure*}[t!]
     \centering
     \begin{subfigure}[b]{0.3333\textwidth}
         \centering
         \includegraphics[width=\textwidth]{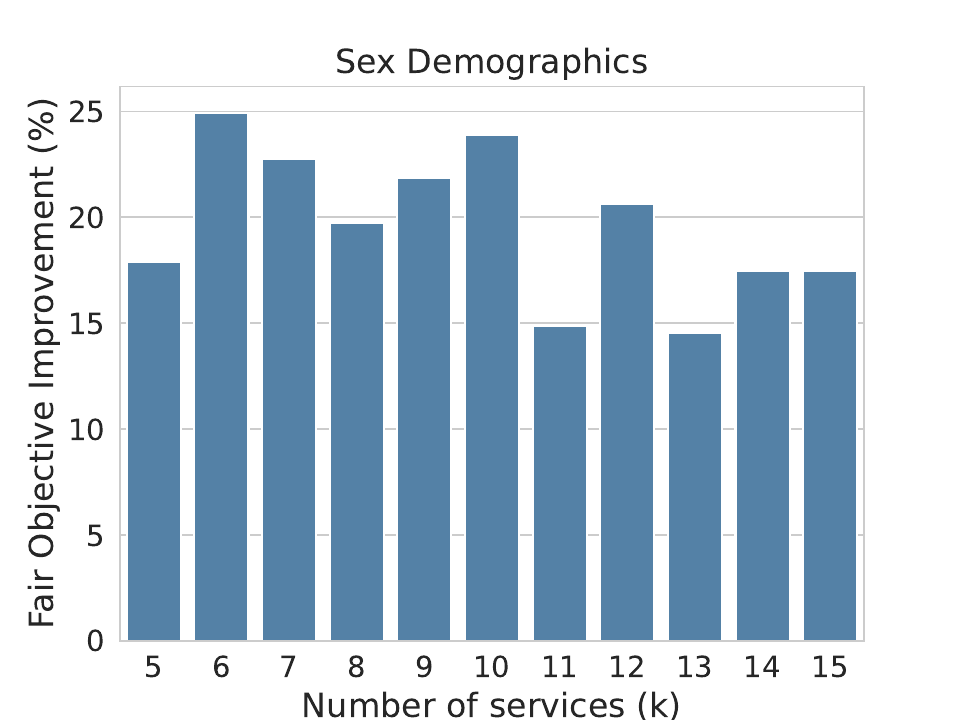}
     \end{subfigure}$\quad$
     \begin{subfigure}[b]{0.3333\textwidth}
         \centering
         \includegraphics[width=\textwidth]{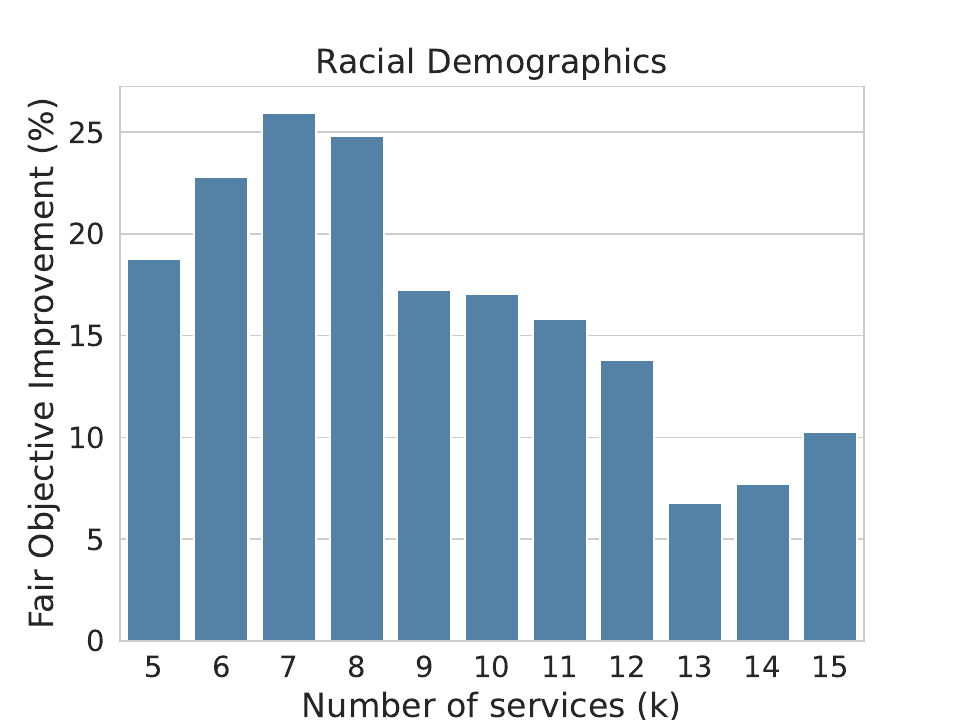}
     \end{subfigure}
     
        \caption{Fair objective improvement for AcQUIre over the baseline across different demographics. We observe that there is atleast 15\% improvement across sex demographics for a wide range of number of services. For racial demographics the improvement is 7-26\%.} 
        % \ljr{Expand this caption to be self contained, what we are seeing, and explain the take away.}}
        \label{fig:fairness_improvement}
\end{figure*}
\begin{figure*}[t!]
     \centering
     \begin{subfigure}[b]{0.32\textwidth}
         \centering
         \includegraphics[width=\textwidth]{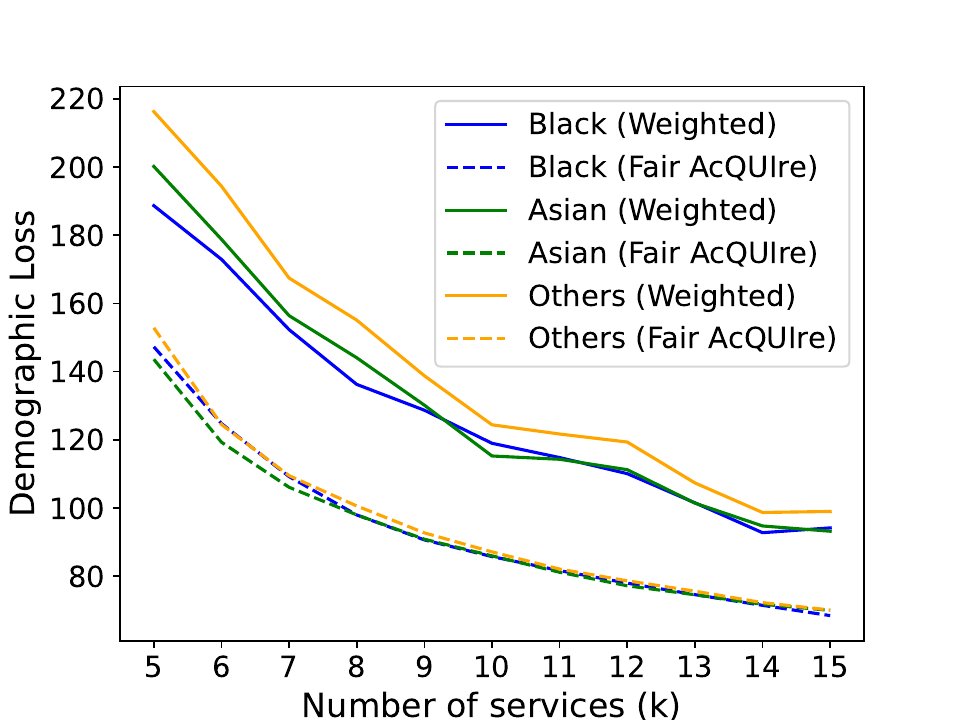}
     \end{subfigure}
     \hfill
     \begin{subfigure}[b]{0.32\textwidth}
         \centering
         \includegraphics[width=\textwidth]{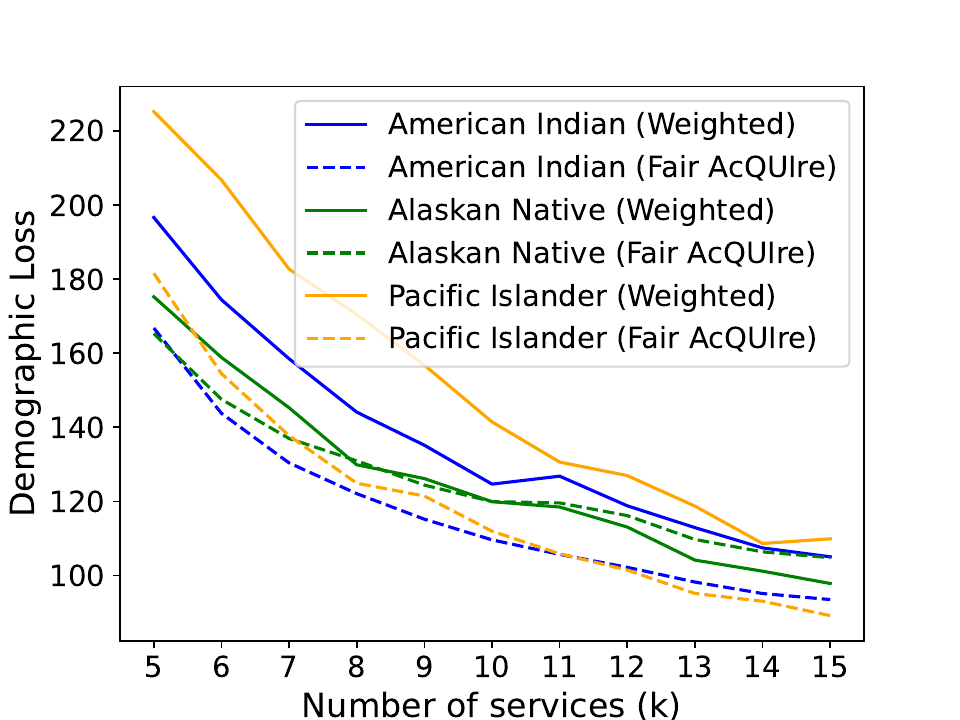}
     \end{subfigure}
     \begin{subfigure}[b]{0.33\textwidth}
         \centering
         \includegraphics[width=\textwidth]{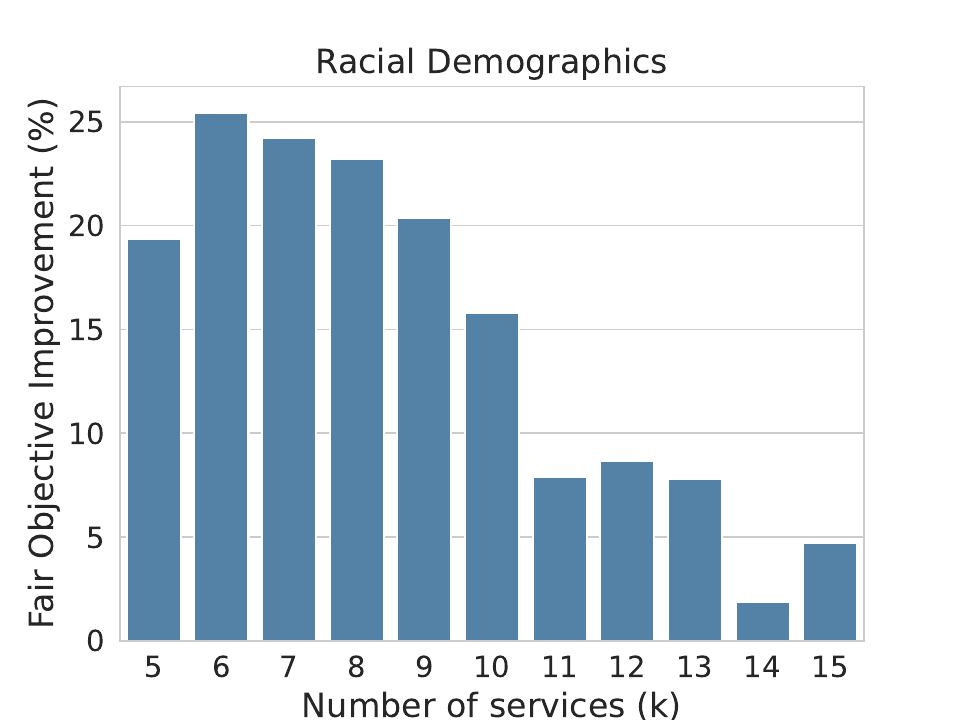}
     \end{subfigure}
     
        \caption{Average losses across different demographic groups for Fair AcQUIre (left,middle). Percentage improvement over baseline (right). We observe that Fair AcQUIre reduces disparity across different groups compared to the baseline. }
        % \ljr{expand this caption to cover all the graphics included and explain the details in a self-contained manner. The third figure you never even reference in the text either so you should explain what it is we are looking at and what the take away is}}
        \label{fig:fairness_new}
\end{figure*}
% \begin{figure}[t!]
%     \centering
%     \includegraphics[width=0.5\textwidth]{new_baselines.pdf}
%     \caption{Average Total Loss across different baselines. We observe that while the greedy algorithm does when the number of services is small, its performance starts degrading as the number of services grows. This is beacuse the greedy algorithm is susceptible to pick outliers. AcQUIre beats both the baselines across a large range of number of services.}
%     % \ljr{expand this caption to explain the results and be self contained}}
%     \label{fig:additionalbaseline}
% \end{figure}
\section{Experiment Details}\label{sec:add_exp}
\textbf{Census Dataset.} The categorical features (schooling, marital status, migration status, citizenship) were first  converted to one hot vectors and the continuous features (age, income) were scaled appropriately for better conditioning. We performed singular value decomposition on the features and retained the top ten components. The scores were taken as the log transform of the daily commute time in minutes. To form subpopulations, we split users based on their Public Use Microdata Area codes (zip code) and ensure each subpopulation belongs to one of the demographic groups considered. The percentage improvements of our algorithm are plotted in Figure~\ref{fig:fairness_improvement}.

In Figure~\ref{fig:fairness_new} we plot the average losses on the individual demographic groups. We benchmark against the weighted random baseline and note that Fair AcQUIre improves not only the fair objective value but the loss for every demographic group. 
% We also run Fair AcQUIre (Algorithm~\ref{alg:fair_initialize}) on the Census experiment where the demographic identities are known to the provider. As a baseline we used a weighted sampling of users where the weights are inversely proportional to the demographic sizes. This baseline ensures the expected number of users sampled per demographic is equal. The results are reported in Figure~\ref{fig:fairness_new}.

% \textbf{Additional Baseline.} We consider an additional baseline for initialization process. The greedy algorithm at every iteration picks the user with the largest loss, queries the picked user's preference and sets the new service to this preference. We find that while the algorithm performs well when the number of services $k$ is small, however as $k$ increases the uniform sampling starts performing better than greedy (see Figure~\ref{fig:additionalbaseline}). This is because a greedy algorithm is susceptible to pick outliers, particularly when $k \ll n$ and these outliers don't generalize well to the average population. 

\textbf{Movie Recommendation Dataset.} We use Surprise (a Python toolkit \citep{Hug2020}) to perform our experiments. We split the total 10 million ratings into top (a) 200 movies, and (b) all other movies. We use the inbuilt nonnegative matrix factorization  function of Surprise on (b) to get user and item embeddings. We cluster the users into 1000 subpopulations by running $k$-means on the obtained user embeddings we get. We evaluate algorithms for this experiment on the held out set (a) of the top 200 movies.

\begin{figure*}[h]
     \centering
     \begin{subfigure}[b]{0.32\textwidth}
         \centering
         \includegraphics[width=\textwidth]{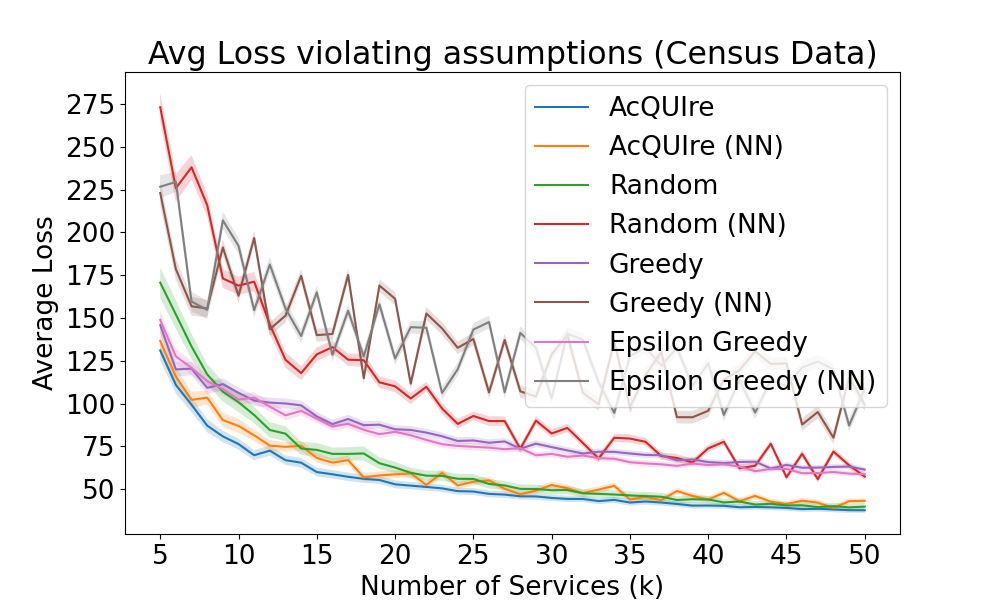}
     \end{subfigure}
     \begin{subfigure}[b]{0.32\textwidth}
         \centering
         \includegraphics[width=\textwidth]{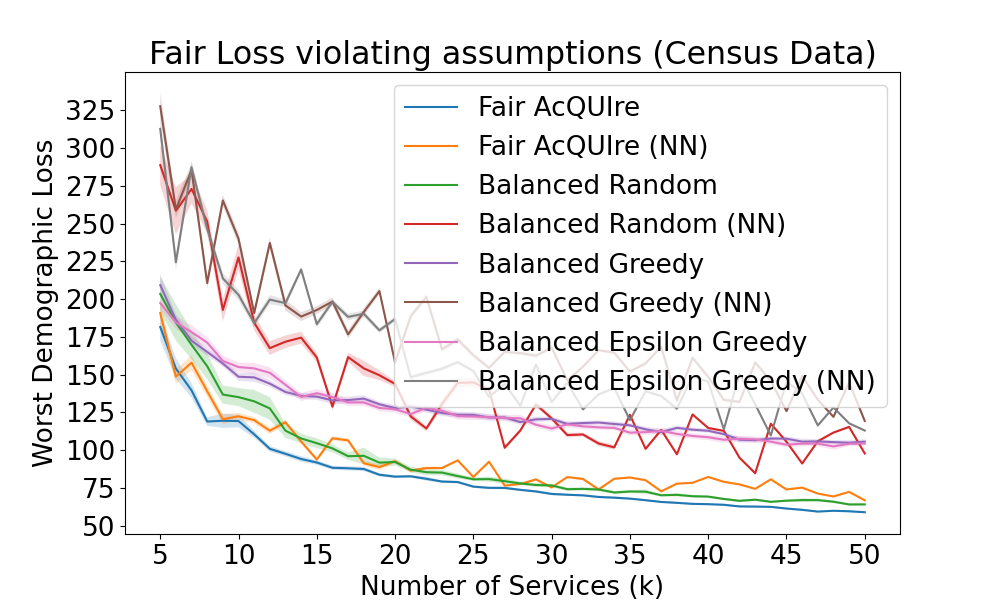}
     \end{subfigure}
     \begin{subfigure}[b]{0.32\textwidth}
         \centering
         \includegraphics[width=\textwidth]{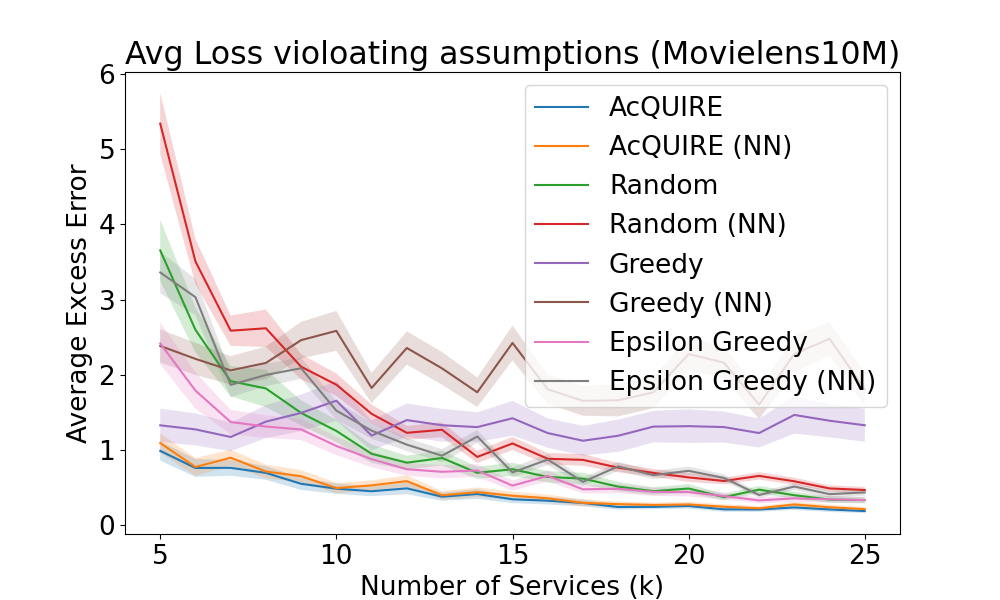}
     \end{subfigure}
        \caption{ We evaluate the performance of initialization methods when Assumptions 2.1 and 2.2 are violated. "AcQUIre (NN)" refers to using AcQUIre with a Neural Network (NN) to predict scores. This terminology is consistently applied to all other baselines as well. Notably,  the performance of AcQUIre doesn't degrade when using Neural Network to predict the scores, thereby demonstrating its robustness to violating the assumptions made.} 
        \label{fig:ablation}
\end{figure*}

\textbf{Ablation.} We use a 2 layer Neural Network with ReLU activations that takes as input the user features and outputs their score. We still use the standard squared prediction error. However note that in this modeling scenario, the loss no longer satisfies our assumptions in the parameters of the neural network. Given a user’s features and true score, since there are no unique minimizers, we run gradient descent to compute a local minimizer and then use this trained neural network as a service to predict other user’s scores. We run AcQUIre and other baselines under this modeling and report our results in Figure~\ref{fig:ablation}. We find that the performance of AcQUIre even when violating assumptions is almost similar to using AcQUIre under modeling which satisfies assumptions.

Additionally, we would like to emphasize that the implementation of our algorithm itself does not rely on these assumptions. It only requires the loss values to be observed. Therefore, one can model very complex precision models and only supply the loss values of these models to our algorithm.

\newpage
\newpage 

\section*{NeurIPS Paper Checklist}

\begin{enumerate}

\item {\bf Claims}
    \item[] Question: Do the main claims made in the abstract and introduction accurately reflect the paper's contributions and scope?
    \item[] Answer: \answerYes % Replace by \answerYes{}, \answerNo{}, or \answerNA{}.
    \item[] Justification: We abstract out the problem statement, challenges, and novelty in solution ideas in the abstract.

\item {\bf Limitations}
    \item[] Question: Does the paper discuss the limitations of the work performed by the authors?
    \item[] Answer: \answerYes % Replace by \answerYes{}, \answerNo{}, or \answerNA{}.
    \item[] Justification: Yes, we clearly state all the assumptions that the theoretical results rely on. We also briefly discuss open directions which were not covered in this paper.

\item {\bf Theory Assumptions and Proofs}
    \item[] Question: For each theoretical result, does the paper provide the full set of assumptions and a complete (and correct) proof?
    \item[] Answer: \answerYes % Replace by \answerYes{}, \answerNo{}, or \answerNA{}.
    \item[] Justification: Proof sketches are in the main paper, with full proofs in the appendix.

    \item {\bf Experimental Result Reproducibility}
    \item[] Question: Does the paper fully disclose all the information needed to reproduce the main experimental results of the paper to the extent that it affects the main claims and/or conclusions of the paper (regardless of whether the code and data are provided or not)?
    \item[] Answer: \answerYes % Replace by \answerYes{}, \answerNo{}, or \answerNA{}.
    \item[] Justification: We state experimental setup and details, and provide anonymous link to our code base and links to the two publicly available datasets that we used in the experiments. 

\item {\bf Open access to data and code}
    \item[] Question: Does the paper provide open access to the data and code, with sufficient instructions to faithfully reproduce the main experimental results, as described in supplemental material?
    \item[] Answer: \answerYes % Replace by \answerYes{}, \answerNo{}, or \answerNA{}.
    \item[] Justification: Yes, we provide both code link and upload codebase in the supplemental.

\item {\bf Experimental Setting/Details}
    \item[] Question: Does the paper specify all the training and test details (e.g., data splits, hyperparameters, how they were chosen, type of optimizer, etc.) necessary to understand the results?
    \item[] Answer: \answerYes % Replace by \answerYes{}, \answerNo{}, or \answerNA{}.
    \item[] Justification: The focus of the paper is not experimental, but experiments are used to illustrate our algorithmic and theoretical contributions. All details are specified in the experiments section and the appendix.

\item {\bf Experiment Statistical Significance}
    \item[] Question: Does the paper report error bars suitably and correctly defined or other appropriate information about the statistical significance of the experiments?
    \item[] Answer: \answerYes % Replace by \answerYes{}, \answerNo{}, or \answerNA{}.
    \item[] Justification: We report number of trials and error bars.

\item {\bf Experiments Compute Resources}
    \item[] Question: For each experiment, does the paper provide sufficient information on the computer resources (type of compute workers, memory, time of execution) needed to reproduce the experiments?
    \item[] Answer: \answerYes % Replace by \answerYes{}, \answerNo{}, or \answerNA{}.
    \item[] Justification: All our experiments can be run on personal devices. 
    %given the computational efficieincy of our algorithm.
    
\item {\bf Code Of Ethics}
    \item[] Question: Does the research conducted in the paper conform, in every respect, with the NeurIPS Code of Ethics \url{https://neurips.cc/public/EthicsGuidelines}?
    \item[] Answer: \answerYes % Replace by \answerYes{}, \answerNo{}, or \answerNA{}.
    \item[] Justification: We have read the code of ethics and verified we align with it.

\item {\bf Broader Impacts}
    \item[] Question: Does the paper discuss both potential positive societal impacts and negative societal impacts of the work performed?
    \item[] Answer: \answerYes % Replace by \answerYes{}, \answerNo{}, or \answerNA{}.
    \item[] Justification: While it is not the main focus of the paper, our paper has a devoted section that considers the \emph{fair version} of the total loss, and discusses the fairness aspects of the proposed methods. 
    
\item {\bf Safeguards}
    \item[] Question: Does the paper describe safeguards that have been put in place for responsible release of data or models that have a high risk for misuse (e.g., pretrained language models, image generators, or scraped datasets)?
    \item[] Answer: \answerNA % Replace by \answerYes{}, \answerNo{}, or \answerNA{}.
    \item[] Justification: The paper is mostly theoretical in nature and poses no such risk.

\item {\bf Licenses for existing assets}
    \item[] Question: Are the creators or original owners of assets (e.g., code, data, models), used in the paper, properly credited and are the license and terms of use explicitly mentioned and properly respected?
    \item[] Answer: \answerYes % Replace by \answerYes{}, \answerNo{}, or \answerNA{}.
    \item[] Justification: We cite the dataset papers.

\item {\bf New Assets}
    \item[] Question: Are new assets introduced in the paper well documented and is the documentation provided alongside the assets?
    \item[] Answer: \answerNA % Replace by \answerYes{}, \answerNo{}, or \answerNA{}.
    \item[] Justification: This paper is theoretical in nature.

\item {\bf Crowdsourcing and Research with Human Subjects}
    \item[] Question: For crowdsourcing experiments and research with human subjects, does the paper include the full text of instructions given to participants and screenshots, if applicable, as well as details about compensation (if any)? 
    \item[] Answer: \answerNA % Replace by \answerYes{}, \answerNo{}, or \answerNA{}.
    \item[] Justification: No human subjects were needed for this paper.

\item {\bf Institutional Review Board (IRB) Approvals or Equivalent for Research with Human Subjects}
    \item[] Question: Does the paper describe potential risks incurred by study participants, whether such risks were disclosed to the subjects, and whether Institutional Review Board (IRB) approvals (or an equivalent approval/review based on the requirements of your country or institution) were obtained?
    \item[] Answer: \answerNA % Replace by \answerYes{}, \answerNo{}, or \answerNA{}.
    \item[] Justification: No.

\end{enumerate}

\end{document}